\documentclass[11pt]{article} 
\usepackage{fullpage}

\usepackage{amsmath,amsfonts,bm}

\def\eqref#1{equation~\ref{#1}}

\def\1{\bm{1}}

\def\rvy{{\mathbf{y}}}

\def\vsigma{\boldsymbol{\sigma}}
\def\vdelta{{\boldsymbol{\delta}}}
\def\vpi{{\boldsymbol{\pi}}}
\def\vvarepsilon{{\boldsymbol{\varepsilon}}}
\def\va{{\mathbf{a}}}

\def\vu{{\mathbf{u}}}
\def\vv{{\mathbf{v}}}

\def\vx{{\mathbf{x}}}
\def\vy{{\mathbf{y}}}
\def\vz{{\mathbf{z}}}

\def\eva{{a}}

\def\evu{{u}}
\def\evv{{v}}

\def\mA{{\mathbf{A}}}
\def\mB{{\mathbf{B}}}
\def\mC{{\mathbf{C}}}
\def\mD{{\mathbf{D}}}

\def\mI{{\mathbf{I}}}

\def\mP{{\mathbf{P}}}

\def\mU{{\mathbf{U}}}

\def\mX{{\mathbf{X}}}
\def\mY{{\mathbf{Y}}}
\def\mZ{{\mathbf{Z}}}

\DeclareMathAlphabet{\mathsfit}{\encodingdefault}{\sfdefault}{m}{sl}
\SetMathAlphabet{\mathsfit}{bold}{\encodingdefault}{\sfdefault}{bx}{n}

\def\gA{{\mathcal{A}}}

\def\gD{{\mathcal{D}}}
\def\gE{{\mathcal{E}}}
\def\gF{{\mathcal{F}}}
\def\gG{{\mathcal{G}}}
\def\gH{{\mathcal{H}}}
\def\gI{{\mathcal{I}}}

\def\gL{{\mathcal{L}}}

\def\gS{{\mathcal{S}}}

\def\gV{{\mathcal{V}}}

\def\gX{{\mathcal{X}}}
\def\gY{{\mathcal{Y}}}

\def\sI{{\mathbb{I}}}

\def\sN{{\mathbb{N}}}

\def\emA{{A}}

\def\emD{{D}}

\def\emP{{P}}

\def\od{{\mathrm{d}}}

\newcommand{\E}{\mathbb{E}}

\newcommand{\R}{\mathbb{R}}

\DeclareMathOperator*{\argmax}{arg\,max}

\newcommand{\divg}{\mathrm{div}}
\newcommand{\diag}{\mathrm{diag}}

\newcommand{\vect}{\mathrm{vec}}

\usepackage{url}

\usepackage{xcolor}
\usepackage{hyperref}
\definecolor{cite_color}{HTML}{114083}
\definecolor{link_color}{RGB}{153,0,0}  
\definecolor{url_color}{RGB}{153,102,0}
\definecolor{emp_color}{RGB}{0,0,255}
\hypersetup{
	colorlinks,
	citecolor=cite_color,
	linkcolor=link_color,
	urlcolor=url_color}

\usepackage{amsmath}
\usepackage{amssymb}
\usepackage{amsthm}
\usepackage{amscd}
\usepackage{amsfonts}
\usepackage{mathrsfs}
\usepackage{graphicx}
\usepackage{tabularborder}
\usepackage{tablefootnote}
\usepackage{enumerate}
\usepackage{bm}
\usepackage{bbm}
\usepackage{natbib}
\usepackage{multirow}
\usepackage{algorithm}
\usepackage{algorithmic}
\usepackage{fancyhdr}
\usepackage{subcaption}
\usepackage{wrapfig}
\usepackage{framed}
\usepackage{color}
\definecolor{shadecolor}{rgb}{0.94, 0.97, 1.0}

\usepackage{tikz}
\usetikzlibrary{graphs, positioning, quotes, shapes.geometric}

\usepackage[capitalize,noabbrev]{cleveref}
\theoremstyle{plain}
\newtheorem{theorem}{Theorem}[section]
\newtheorem{proposition}[theorem]{Proposition}
\newtheorem{lemma}[theorem]{Lemma}
\newtheorem{corollary}[theorem]{Corollary}

\theoremstyle{definition}
\newtheorem{definition}[theorem]{Definition}
\newtheorem{assumption}{Assumption}
\theoremstyle{remark}
\newtheorem{remark}[theorem]{Remark}

\crefname{section}{Section}{Sections}
\crefname{theorem}{Theorem}{Theorems}
\crefname{lemma}{Lemma}{Lemmas}
\crefname{corollary}{Corollary}{Corollary}
\crefname{equation}{Eq.}{Eq.}
\crefname{proposition}{Proposition}{Propositions}
\crefname{claim}{Claim}{Claims}
\crefname{remark}{Remark}{Remarks}
\crefname{assumption}{Assumption}{Assumptions}
\crefname{definition}{Definition}{Definitions}
\crefname{appendix}{Appendix}{Appendices}
\crefname{algorithm}{Algorithm}{Algorithms}
\crefname{figure}{Figure}{Figures}
\crefname{table}{Table}{Tables}

\newcommand{\appendixtitle}[1]{
    \begin{center}
        \LARGE \bf #1
    \end{center}
}
\newcommand{\gray}[1]{\textcolor{gray}{#1}}

\usepackage{etoc}
\etocdepthtag.toc{mtchapter}
\etocsettagdepth{mtchapter}{subsection}
\etocsettagdepth{mtappendix}{none}

\title{\textbf{Implicit Graph Neural Diffusion Networks: Convergence, Generalization, and Over-Smoothing}}
\author{
  Guoji Fu \quad Mohammed Haroon Dupty \quad Yanfei Dong \quad Lee Wee Sun \\
  School of Computing \\
  National University of Singapore \\
  \textit{\small\{guojifu, dmharoon, dyanfei, leews\}@comp.nus.edu.sg}
}

\begin{document}

\maketitle

\begin{abstract}
    Implicit Graph Neural Networks (GNNs) have achieved significant success in addressing graph learning problems recently.
    However, poorly designed implicit GNN layers may have limited adaptability to learn graph metrics, experience over-smoothing issues, or exhibit suboptimal convergence and generalization properties, potentially hindering their practical performance. 
    To tackle these issues, we introduce a geometric framework for designing implicit graph diffusion layers based on a parameterized graph Laplacian operator. 
    Our framework allows learning the metrics of vertex and edge spaces, as well as the graph diffusion strength from data.
    We show how implicit GNN layers can be viewed as the fixed-point equation of a Dirichlet energy minimization problem and give conditions under which it may suffer from over-smoothing during training (OST) and inference (OSI).
    We further propose a new implicit GNN model to avoid OST and OSI. 
    We establish that with an appropriately chosen hyperparameter greater than the largest eigenvalue of the parameterized graph Laplacian, DIGNN guarantees a unique equilibrium, quick convergence, and strong generalization bounds.
    Our models demonstrate better performance than most implicit and explicit GNN baselines on benchmark datasets for both node and graph classification tasks.
    Code available at \url{https://github.com/guoji-fu/DIGNN}.
\end{abstract}

\section{Introduction}\label{sec:intro}
Graph Neural Networks (GNNs) have achieved great success across various domains, including social media, natural language processing, computational biology, etc.
Existing GNN architectures include explicit and implicit models.
Explicit GNN models, such as graph convolutional networks~\citep{ChebyNet-NIPS2016,GCN-ICLR2017,GraphSage-NIPS2017}, graph attention networks~\citep{GAT-ICLR2018,GANN-aXiv2018}, message passing neural networks~\citep{MPNN-ICML2017,APPNP-ICLR2019,GPRGNN-ICLR2021,pGNN-ICML2022}, learn node representations by iteratively propagating and transforming node features over the graph in each layer.
In contrast, implicit GNN models, which is the focus of this work, adopt an implicit layer~\citep{DEQ-NIPS2019,IDL-aXiv2019} to replace deeply stacked explicit propagation layers.
The implicit layer solves a fixed point equilibrium equation using root-finding methods and is equivalent to running infinite propagation steps~\citep{IGNN-NIPS2020,GIND-ICML2022}.
The obtained equilibrium can then be used as node representations, which contain information from infinite-hop neighbors.

The performance of implicit GNNs depends significantly on the adopted implicit layer. 
Simply generalizing explicit GNN layers to implicit layers could inherit the over-smoothing issue~\citep{GNNExpress-ICLR2020,DGC-NIPS2021,GRAND++-ICLR2022,GIND-ICML2022}.
Besides, the underlying graph metric has been shown to have a profound connection with the performance of GNNs in heterophilic settings and their over-smoothing behaviours~\citep{Sheaf-NIPS2022}.
However, existing implicit GNNs, e.g., \citet{CGNN-ICML2020,IGNN-NIPS2020,EIGNN-NIPS2021,GIND-ICML2022,GRAND-ICML2021,GRAND++-ICLR2022}, have primarily concentrated on learning the diffusion strength without explicitly engaging in learning the metrics of vertex and edge spaces, which may limit their adaptability for learning the graph metrics.
Furthermore, reliability, including both the \textit{convergence} of implicit layers~\citep{CGS-ICLR2022} and the \textit{generalization} ability to unseen data, stands as a vital criterion for implicit GNNs.  
While numerous studies, e.g., \citet{gener/nn/ScarselliTH18,gener/kdd/VermaZ19,gener/icml/GargJJ20,gener/nips/OonoS20,gener/iclr/LiaoUZ21,gener/ijon/ZhouW21,gener/nips/EsserVG21,gener/icml/YehudaiFMCM21,gener/icml/tang23}, have focused on the generalization abilities of explicit GNNs, this aspect for implicit GNNs is relatively overlooked in existing implicit GNN literature, where the emphasis has been on ensuring convergence properties.

To confront the above challenges, we introduce a geometric framework to equip implicit GNNs with the ability to learn the graph metrics while avoiding over-smoothing and guaranteeing reliable convergence and generalization properties.
Specifically, we define \textit{parameterized vertex and edge Hilbert spaces} to learn the graph metrics and a \textit{parameterized graph gradient} operator to learn the graph diffusion strength directly from data. 
They further facilitate the construction of a \textit{parameterized graph Laplacian} operator whose spectrum is crucial in analyzing over-smoothing, convergence, and generalization behaviours of our models.

We define the \textit{parameterized Dirichlet energy} as the squared norm of the parameterized graph gradient flows on the edge Hilbert space, providing a measure of smoothness over a graph. 
We show that the implicit graph diffusion can be viewed as a fixed-point equilibrium equation of an optimization problem that aims to minimize the Dirichlet energy to ensure smoothness, subject to constraints, e.g. label constraints, at some vertices~\citep{SSL-NIPS2004,SSL-COLT2004}. 
We identify that if our graph parameterizations are not formulated as functions of node features, the implicit graph diffusion may suffer from the same issue as characterized in some explicit GNNs (see e.g., \citet{OS-AAAI2018,DGC-NIPS2021}), which we term as \textit{over-smoothing during training (OST)}: the output node representations are independent of the input node features during training.
Furthermore, for tasks without node label constraints during inference, e.g., graph classification, the graph representations obtained by directly minimizing the Dirichlet energy may converge to a constant function over all nodes, regardless of input features.
We call this phenomenon, which has also been observed in a graph neural diffusion model by \citet{GRAND++-ICLR2022}, as \textit{over-smoothing during inference (OSI)}.

Based on our framework, we design a new implicit GNN model, Dirichlet Implicit Graph Neural Network (DIGNN), whose implicit graph diffusion layer is derived from a parameterized Dirichlet energy minimization problem with constraints on input node features. 
The incorporation of parameterizations and constraints related to node features allows DIGNN to learn the graph metrics while avoiding OST and OSI.
We derive the graph diffusion convergence results and transductive generalization bounds for DIGNN, demonstrating that with an appropriately chosen hyperparameter greater than the \textit{largest eigenvalue} of the parameterized graph Laplacian, DIGNN has a unique equilibrium and guarantees quick convergence rates and strong generalization bounds.
Moreover, our findings reveal a significant link between convergence and generalization properties for implicit GNNs, indicating that superior generalization capability requires fast convergence for implicit graph diffusion.

We validate the effectiveness of our model and theoretical findings on a variety of benchmark datasets for node and graph classification tasks. 
The experimental results show that our model obtains better performance than explicit and implicit GNN baselines for both types of tasks in most cases.

We summarize our contributions below:
\begin{itemize}
    \item We introduce a framework to enable implicit GNNs to learn the graph metrics directly from data while avoiding over-smoothing and guaranteeing reliable convergence and generalization properties.

    \item We show that the implicit graph diffusion can be viewed as the fixed-point equation of a Dirichlet energy minimization problem and give conditions under which it may suffer from over-smoothing.

    \item We propose a new implicit GNN model and derive new convergence and generalization results. Our theory reveals that superior generalization capability requires fast convergence for implicit graph diffusion.

    \item Our model empirically performs well on benchmark datasets for both node and graph classification tasks.
\end{itemize}
We defer all proofs to the Appendix.

\section{Related Work}\label{sec:related}
Here we briefly discuss the related work and defer a more detailed discussion to \cref{sec:app:related}.

\textbf{Implicit GNNs and graph neural diffusions.} 
To capture long-range dependencies in graphs, implicit GNNs define their outputs as solutions to fixed-point equations, which are equivalent to running infinite-depth GNNs.
A series of implicit GNN architectures, e,g., \citet{GNN-TNN2009,IGNN-NIPS2020,EIGNN-NIPS2021,CGS-ICLR2022,GIND-ICML2022,MGNNI-NIPS2022,Opt-EquiPAMI2023,IGNN-icml/BakerWHW23}, have been proposed where they have adopted different propagation schemes to design their implicit layers.
On the other hand, continuous-depth GNNs, such as \citet{CGNN-ICML2020,GRAND-ICML2021,PDE-GCN-NIPS2021,Beltrami-NIPS2021,GRAFF-arXiv2022,Sheaf-NIPS2022,GRAND++-ICLR2022}, use graph neural diffusion processes to capture long-range dependencies.
Their architectures are formulated through the discretization of neural differential equations~\citep{NDE-NIPS2018,NDE-arXiv2022} on graphs. 
However, existing work has mainly focused on learning the diffusion strength without explicitly engaging in learning the metrics of vertex and edge spaces.
Additionally, the generalization behaviors of implicit GNNs have not been studied. The primary focus in the literature has so far been on ensuring convergence.

\textbf{Theoretical analysis of GNNs.} Numerous efforts have been made in theoretically analyzing the behaviours of explicit GNNs, focusing on over-smoothing issues~\citep{OS-AAAI2018,OS/iclr/ZhaoA20,OS-CIKM2021,GNNExpress-ICLR2020,OS-NIPS2022} and establishing generalization bounds~\citep{gener/nn/ScarselliTH18,gener/kdd/VermaZ19,gener/icml/GargJJ20,gener/nips/OonoS20,gener/iclr/LiaoUZ21,gener/ijon/ZhouW21,gener/nips/EsserVG21,gener/icml/tang23}.
In this work, focusing on implicit GNNs, we characterize the conditions that lead to over-smoothing issues and derive new convergence and generalization results.
\section{Preliminaries and Background}\label{sec:pre}
We use $\R_+ = \{x \in \R \mid x \geq 0\}$ to denote the space of non-negative real values and $\R_+^* = \{x \in \R \mid x > 0\}$ to denote the space of positive real values.
We denote scalars by lower- or upper-case letters and vectors and matrices by lower- and upper-case boldface letters, respectively. Given a matrix $\mA$, $\mA_{i,:}$ is its $i$-th row vector and $\mA_{:,j}$ is its $j$-column vector, we denote its transpose by $\mA^\top$ and its Hadamard product with another matrix $\mB$ by $\mA \odot \mB$, i.e., the element-wise multiplication of $\mA$ and $\mB$. We use $\|\mA\|$ to denote the spectral norm of a 
matrix $\mA$. A set $[1, 2, \dots, n]$ is denoted as $[n]$.
Let $\gG = (\gV, \gE, \mA)$ be an undirected graph where $\gV = [n]$ denotes the set of vertices, $\gE \subseteq \gV \times \gV$ denotes the set of edges, $\mA \in \R_+^{n \times n}$ is the adjacency matrix and $\emA_{i,j} = \emA_{j,i}, \emA_{i,j} > 0$ for $[i, j] \in \gE$, $\emA_{i,j} = 0$ otherwise. $\mD \in \R_+^{n \times n} = \diag(\emD_1, \dots, \emD_n)$ denotes the diagonal degree matrix with $\emD_i = \sum_j\emA_{i,j}, i = 1, \dots, n$.
The meaning of other notations can be inferred from the context.

\subsection{Hilbert space of functions on vertices and edges}
We define two inner products as the graph metrics to model the Hilbert spaces on vertices and edges.

\begin{definition}[Inner Product on the Vertex Space]\label{def:def2}
    Given a graph $\gG$ and any functions $f: \gV \mapsto \R, g: \gV \mapsto \R$, the inner product on the function space $\R^\gV$ on $\gV$ is defined as
    \begin{equation}
        \langle f, g \rangle_\gV = \sum_{i \in \gV}f(i)g(i)\chi(i),
    \end{equation}
    where $\chi: \gV \mapsto \R_+$ is a positive real-valued function.
\end{definition}

\begin{definition}[Inner Product on the Edge Space]\label{def:def3}
    Given a graph $\gG$ and any functions $F: \gE \mapsto \R, G: \gE \mapsto \R$, the inner product on the function space $\R^\gE$ on $\gE$ is defined as
    \begin{equation}
        \langle F, G \rangle_\gE = \frac{1}{2}\sum_{[i, j] \in \gE}F([i, j])G([i, j])\phi([i,j]),
    \end{equation}
    where $\phi: \gE \mapsto \R_+$ is a positive real-valued function.
\end{definition}
The inner products on vertex and edge spaces induce two norms $\|\cdot\|_\gV = \sqrt{\langle \cdot, \cdot \rangle_\gV}$ and $\|\cdot\|_\gE = \sqrt{\langle \cdot, \cdot \rangle_\gE}$, respectively. We denote by $\gH(\gV, \chi) = (\R^\gV, \langle \cdot, \cdot \rangle_\gV)$ and $\gH(\gE, \phi) = (\R^\gE, \langle \cdot, \cdot \rangle_\gE)$ the corresponding Hilbert spaces.

\begin{remark}[The cases for vector functions]
    For vector functions $f, g: \gV \mapsto \R^d$, the inner product on the vertex space is given by $\langle f, g \rangle_\gV = \sum_{i \in \gV}\langle f(i), g(i) \rangle\chi(i)$. Similarly, for vector functions $F, G: \gE \mapsto \R^c$, the inner product on the edge space is given by $\langle F, G \rangle_\gE = \frac{1}{2}\sum_{[i, j] \in \gE}\langle F([i, j]), G([i, j]) \rangle\phi([i,j])$.
\end{remark}

\subsection{Parameterized graph Laplacian operator}
We start by defining the parameterized graph gradient and divergence operators. Subsequently, these operators are used to construct the parameterized graph Laplacian operator.

\begin{definition}[Parameterized Graph Gradient]\label{def:def4}
    Given a graph $\gG$ and a function $f: \gV \mapsto \R$, the parameterized graph gradient $\nabla: \gH(\gV, \chi) \mapsto \gH(\gE, \phi)$ is defined as:
    \begin{equation}
        (\nabla f)([i, j]) = \varphi([i, j])(f(j) - f(i)), \forall [i, j] \in \gE,
    \end{equation}
    where $\varphi: \gE \mapsto \R_+^*$ is a positive real-valued function.
\end{definition}
Intuitively, $(\nabla f)([i, j])$ measures the variation of $f$ on an edge $[i, j] \in \gE$. We define the parameterized graph divergence as the adjoint of the parameterized graph gradient:

\begin{definition}[Parameterized Graph Divergence]\label{def:def5}
    For any functions $f: \gV \mapsto \R, g: \E \mapsto \R$, the parameterized graph divergence $\divg: \gH(\gE, \phi) \mapsto \gH(\gV, \chi)$ is defined by
    \begin{equation}
        \langle \nabla f, g \rangle_\gE = \langle f, -\divg g \rangle_\gV.
    \end{equation}
\end{definition}

\begin{remark}
    Note that $f$ in \cref{def:def4} and $g$ in \cref{def:def5} can be some vector functions $f: \gV \mapsto \R^d, g: \gE \mapsto \R^d$ for some $d \in \sN$. Extending the definitions of the gradient operator and its adjoint and the graph Laplacian to vector functions is straightforward by using the inner products on vertex and edge spaces for vector functions.   
\end{remark}

\begin{lemma}\label[lemma]{thrm:lem1}
    Given an undirected graph $\gG = (\gV, \gE)$ and a function $g: \gE \mapsto \R$, the parameterized graph divergence $\divg: \gH(\gE, \phi) \mapsto \gH(\gV, \chi)$ is explicitly given by
    \begin{equation*}
        (\divg g)(i) = \frac{1}{2\chi(i)}\sum_{j=1}^n\varphi([i, j])\phi([i, j])\left(g([i, j]) - g([j, i])\right).
    \end{equation*}
\end{lemma}
Intuitively, $(\divg g)(i)$ measures the inflows of $g$ towards vertex $i$ and its outflows from $i$.

\begin{definition}[Parameterized Graph Laplacian Operator]\label{def:graph-lap}
    Given positive real-valued functions $\chi: \gV \mapsto \R_+, \phi: \gE \mapsto \R_+, \varphi: \gE \mapsto \R_+^*$ ($\Phi := \chi \cup \phi \cup \varphi$), the parameterized graph Laplacian $\Delta_\Phi: \gH(\gV, \chi) \mapsto \gH(\gV, \chi)$ is defined as
    \begin{equation}
        \Delta_\Phi = -\divg\nabla.
    \end{equation}
\end{definition}

\cref{thrm:lem2} gives the explicit expression of $\Delta_\Phi$: 
\begin{lemma}\label[lemma]{thrm:lem2}
    Given an undirected graph $\gG = (\gV, \gE)$ and a function $f: \gV \mapsto \R$, we have for all $i \in \gV$,
    \begin{equation}\label{eq:param-lap}
        (\Delta_\Phi f)(i) = \frac{1}{\chi(i)}\sum_{j=1}^n\varphi([i, j])^2\phi([i, j])\left(f(i) - f(j)\right).
    \end{equation}
\end{lemma}
We also demonstrate that for any positive real-valued functions $\chi, \phi, \varphi$, the graph Laplacian is \textit{self-adjoint} and \textit{positive semi-definite} by \cref{thrm:prop1} in \cref{app:thrm:prop1}.

\begin{remark}[Compare to canonical Laplacians]
    Canonical graph Laplacian operators, such as unnormalized Laplacian, random walk Laplacian, and normalized Laplacian are special cases in our definition, whose corresponding Hilbert spaces on vertices and edges and graph gradient operator are fixed as summarized in \cref{tab:tab1} in the Appendix. We provide a detailed discussion on these operators in \cref{app:lap}.
\end{remark}

\section{Implicit Graph Diffusions: Convergence and Over-Smoothing Issues}\label{sec:analysis}

\subsection{Parameterized Dirichlet energy}

We define the parameterized Dirichlet energy as the square norm of the parameterized graph gradient flows on the edge Hilbert space (see \cref{app:der:Dirichlet} for detailed derivation): 
\begin{equation}
    \gS(f) = \frac{1}{2}\sum_{i=1}^n\sum_{j=1}^n\varphi([i, j])^2\phi([i, j])\|f(j) - f(i)\|^2.
\end{equation}
$\gS(f)$ measures the variation of $f$ over the whole graph $\gG$ and lower energy graphs are smoother. We have $\left.\frac{\od\gS(f)}{\od f}\right|_i = 2\chi(i)(\Delta_\Phi f)(i)$ by \cref{thrm:lem3} as given in \cref{app:thrm:lem3}.

\subsection{Dirichlet energy minimization}\label{sec:analysis-obj}
Consider an undirected graph $\gG = (\gV, \gE, \mA)$ with node features $\mX = (\vx_1, \dots, \vx_n)^\top \in \R^{n \times d}$, assume that we have a subset of constrained nodes $\gI \subseteq \gV$ where each node $i \in \gI$ is provided with a target vector $\rvy(i) \in \R^c$. 
We aim to learn a function $f = (f(1), \dots, f(n))^\top \in \R^{n \times c}$ to predict the value of all the nodes.
To this end, a natural formulation would be to minimize the Dirichlet energy to ensure smoothness, subject to the function being close to the targets at the constrained nodes:
\begin{equation}\label{eq:Dirichlet}
    \inf_f\gL(f) = \inf_f\gS(f) + \mu\sum_{i \in \gI}\|f(i) - \rvy(i)\|_\gV^2, 
\end{equation}
where the trade-off is controlled by the hyperparameter $\mu$. 
As will be shown in \cref{sec:model-bound}, $\mu$ also significantly influences the convergence and generalization properties of our models.
If $i \in \gI$, let $\delta_i = 1, \rvy'(i) = \rvy(i)$ and otherwise let $\delta_i = 0, \rvy'(i) = \mathbf{0}$ .
Denote by $\vdelta := (\delta_1, \dots, \delta_n) \in \R^n, \mY' := (\rvy'(1), \dots, \rvy'(n))^\top \in \R^{n \times c}$. 
Define $\widehat{\mD} := \diag(\widehat{\emD}_1, \dots, \widehat{\emD}_n)$ with $\widehat{\emD}_i := \sum_k\varphi(i, k)^2\phi([i, k])$, $\mP \in \R^{n \times n}$ with $\emP_{i,j} := \varphi([i, j])^2\phi([i, j]) / \widehat{\emD}_i$, $\mC \in \R^{n \times n}$ with $\mC_{i,:} := (1 - \delta_i)\mP_{i,:} - \delta_i\frac{1}{\mu}\Delta_{i,:}$.

The solution of \cref{eq:Dirichlet} satisfies the fixed-point equilibrium equation (see \cref{app:der:equi} for detailed derivation):
\begin{equation}\label{eq:equi}
    f = \mY' + \mC f.
\end{equation}

\subsection{Convergence analysis}

\begin{lemma}[Convergence and Uniqueness Analysis]\label[lemma]{thrm:lem4}
    Let $\gamma_{\text{max}}$ be the largest singular value of $\mC$. The fixed-point equilibrium equation (\cref{eq:equi}) has a unique solution if $\gamma_{\text{max}} < 1$. We can iterate the equation:
    \begin{equation} 
        f^{(t+1)} = \mY' + \mC f^{(t)}, \quad t = 0, 1, \dots.
    \end{equation}
    to obtain the optimal solution $f^\star = \lim_{t \rightarrow \infty}f^{(t)}$ which is independent of the initial state $f^{(0)}$. 
\end{lemma}

\subsection{Over-smoothing during training and inference}
Specializing the results for other GNN architectures in previous works, e.g., \citet{GNNExpress-ICLR2020,DGC-NIPS2021,GRAND++-ICLR2022,GIND-ICML2022}, we describe the conditions that lead to two types of over-smoothing issues in our fixed-point equation~(\ref{eq:equi}), namely \textit{over-smoothing during training (OST)} and \textit{over-smoothing during inference (OSI)}, as outlined in \cref{thrm:corol1,thrm:lem5} respectively.

\begin{corollary}[OST Condition]\label[corollary]{thrm:corol1}
    Assume that the initial state $f^{(0)}$ at node $i$ is initialized as a function of $\vx_i$. Assume further that $\chi, \phi, \varphi$ are not functions of $\vx_i$, e.g., $\Delta$ is the unnormalized Laplacian or random walk Laplacian. Let $\gamma_{\text{max}}$ be the largest singular value of $\mC$. Then, if $\gamma_{\text{max}} < 1$, the equilibrium of the fixed point equation \cref{eq:equi} is independent of the node features $\vx_i$, i.e. the same equilibrium solution is obtained even if the node features $\vx_i$ are changed.
\end{corollary}

We describe how OST could occur in problem~(\ref{eq:Dirichlet}) where given some labeled nodes, we aim to predict the labels of the remaining nodes. 
In the simplest case, we use $f(i)$ to directly predict the target $y(i)$ by solving \cref{eq:Dirichlet}. 
\cref{thrm:corol1} indicates that the output node representations will be independent of the input node features during \textit{training process} if the parameterizations of vertex and edge Hilbert spaces as well as the graph gradient operator are not defined as functions of node features. 
This implies that node features are not utilized in the learning process of the prediction function, which would be undesirable if they contain useful information for the problem. 
The independence of output from features can be mitigated by making the graph parameterizations to be dependent on node features. 
However, despite parameterizations, over-smoothing may still be present if we use graph representations obtained by directly minimizing the Dirichlet energy for tasks such as graph classification which do not have any constrained nodes during the \textit{inference process}. 
As shown by \cref{thrm:lem5}, when there are no constrained nodes in \cref{eq:Dirichlet}, the equilibrium of \cref{eq:equi} is a constant function over all nodes, which drastically reduces what it can represent.

\begin{lemma}[OSI Condition]\label[lemma]{thrm:lem5}
    When $\gI = \emptyset$, i.e., there is no constrained nodes in \cref{eq:Dirichlet}, we have $\mC = \mP$ and $\rvy'(i) = \mathbf{0}, \forall i \in \gV$. $\mP$ is a Markov matrix and the solution of \cref{eq:Dirichlet} satisfies $f = \mP f$. If the graph $\gG$ is connected and non-bipartite, i,e., the Markov chain corresponding to $\mP$ is irreducible and aperiodic, the equilibrium $f^\star(i)$ for all nodes are identical and not unique, i.e., $f^\star(i) = \vv, \forall i \in \gV$, where $\vv$ is any vector in $\R^c$ and is independent of the labels. Iterating $f^{(t+1)} = \mP f^{(t)}, t = 0, 1, \dots,$ with initial state $f^{(0)} \in \R^{n \times c}$ will converge to
    \begin{equation}\label{eq:osi}
        f^\star(i) = (\lim_{t \rightarrow \infty}\mP^tf^{(0)})(i) = (\vpi f^{(0)})^\top, \quad \forall i \in \gV,
    \end{equation}
    where $\boldsymbol{\pi} = (\widehat{\emD}_1 / \sum_{j}\widehat{\emD}_j, \dots, \widehat{\emD}_n / \sum_{j}\widehat{\emD}_j) \in \R^{1 \times n}$ is the stationary distribution of $\mP$.
\end{lemma}

\begin{remark}[OSI Phenomenon in GRAND Architecture]
    Previous work~\citep{GRAND++-ICLR2022} has derived closely related results on the OSI phenomenon, which we show here for implicit GNNs, for a graph neural diffusion model, i.e., GRAND~\citep{GRAND-ICML2021}. 
\end{remark}

\section{Dirichlet Implicit Graph Neural Networks}\label{sec:model}
Consider an undirected graph $\gG = (\gV, \gE, \mA)$ with node features $\mX \in \R^{n \times d}$.
We present the following assumptions:
\begin{assumption}\label{assump:graph}
    The graph $\gG = (\gV, \gE, \mA)$ is connected.
\end{assumption}

\begin{assumption}\label{assump:embedding}
    For all $i \in \gV$, the node embeddings $\vx_i$ (or $\widetilde{\vx}_i$) are upper-bounded by $\|\vx_i\| \leq c_X$ for some $c_X \in \R_+^*$.
\end{assumption}

\begin{assumption}\label{assump:param}
    The learnable parameters $\Theta_\chi, \Theta_\phi$ are upper-bounded by $\|\Theta_\chi\| \leq c_\chi, \|\Theta_\phi\| \leq c_\phi$ for some $c_\chi, c_\phi \in \R_+^*$.
\end{assumption}

\subsection{Graph neural Laplacian}\label{sec:model:lap}
We design the \textit{graph neural Laplacian} $\Delta_\Phi: \gH(\gV, \chi) \mapsto \gH(\gV, \chi)$ by specifying $\chi, \phi, \varphi$ in the parameterized graph Laplacian (\cref{eq:param-lap}) as below: $\forall i \in \gV$ and $\forall [i, j] \in \gE$,
\begin{align}
    \chi(i) = {} & \emD_i\tanh(\|\Theta_\chi\vx_i\|), \label{eq:eq12} \\
    \phi([i, j]) = {} & \tanh(|(\Theta_\phi\Theta_\chi\vx_i)^\top(\Theta_\phi\Theta_\chi\vx_j)|), \label{eq:eq13} \\
    \varphi([i, j]) = {} & \sqrt{\emA_{i,j}\tanh((\|\Theta_\varphi(\vx_i - \vx_j)\| + \epsilon)^{-1})}, \label{eq:eq14}
\end{align}
where $\epsilon > 0$ is a small value used to avoid dividing zero for neighbor nodes with identical embeddings. 
$\Theta_\chi, \Theta_\phi, \Theta_\varphi$ are matrices of learnable parameters. 
Note that $\phi([i, j]) = \phi([j, i])$ and $\varphi([i, j]) = \varphi([j, i])$ for all $[i, j] \in \gE$.

It is easy to verify that \cref{eq:eq12,eq:eq13,eq:eq14} satisfy the requirements that $\chi, \phi, \varphi$ are strictly positive on $\R_+^*$. 
We adopt $\tanh(\cdot)$ as the nonlinearity as it satisfies the requirements and can be upper-bounded which will be helpful for spectral analysis and in ensuring convergence of the induced implicit layer. 
$\chi$ (\cref{eq:eq12}) and $\phi$ (\cref{eq:eq13}) are used to learn vertex and edge Hilbert spaces, respectively. $\varphi$ (\cref{eq:eq14}) is used to learn the diffusivity of the graph gradient operator. 
Intuitively, the forms of \cref{eq:eq12,eq:eq13,eq:eq14} have the following properties: 
by \cref{eq:eq12}, the impact of node $i$ on the vertex space diminishes when $\|\Theta_\chi\vx_i\|$ is small and vice versa; 
by \cref{eq:eq13}, the impact of an edge $[i, j]$ on the edge space will depend on the learned similarity of node $i$ and $j$ where edges between dissimilar nodes have less influence and vice versa; 
\cref{eq:eq14} focuses on the difference between nodes and would impose larger diffusion weights for similar neighborhood nodes and smaller weights for dissimilar neighbors. 
Notably, here we just provide three feasible choices with additional expected properties for $\chi, \phi, \varphi$. 
In general, other positive real-valued functions can be designed in the graph neural Laplacian to exploit problem-specific properties.

\begin{theorem}[Spectral Range of $\Delta_\Phi$]\label[theorem]{thrm:thrm1}
    Let $\Delta_\Phi$ be the graph neural Laplacian and $\chi, \phi, \varphi$ are given by \cref{eq:eq12,eq:eq13,eq:eq14} respectively, $\lambda_\Phi$ be an eigenvalue associated with the eigenvector $\vu$ of $\Delta_\Phi$. Suppose \cref{assump:graph,assump:embedding,assump:param} hold, then
    \begin{equation}
        0 \leq \lambda_\Phi \leq 2c_\phi^2c_\chi c_X\cosh(c_\chi c_X).
    \end{equation}
\end{theorem}
Theorem~\ref{thrm:thrm1} shows that the eigenvalues will be small if $c_X$,  $\|\Theta_\chi\|, \|\Theta_\phi\|$ are kept small.
As will be shown later, the size of the largest eigenvalue of $\Delta_\Phi$ is essential for the convergence and generalization behaviours of our models.

\subsection{DIGNN architectures}
Drawing upon the preceding theoretical analysis given in \cref{sec:analysis}, we design a novel implicit graph neural network.
We would like the output of the implicit GNN to depend on the features $\mX$; when it does not, we view the problem as over-smoothing. 
A straightforward way to guarantee that the output of the implicit GNN depends on $\mX$ is to ensure that each node is constrained and the target constraint $\rvy(i)$ is a function of $\vx_i$. 
Further, by adjusting the hyperparameter $\mu$, we can control the strength of the constraints, allowing a tradeoff between the exploitation of node features and smoothing. 
We incorporate this idea in the design of our \textit{Dirichlet Implicit Graph Neural Network}, termed DIGNN.
The architecture of DIGNN is given as follows:
\begin{align}
    \widetilde{\mX} = {} & h_{\Theta^{(1)}}(\mA, \mX) \label{eq:model-in} \\
    \mZ = {} & \widetilde{\mX} - \frac{1}{\mu}\Delta\mZ, \label{eq:diff}\\
    \widehat{\vy} = {} & h_{\Theta^{(2)}}(\mZ) \label{eq:model-out}.
\end{align}
The first layer~\cref{eq:model-in} serves as a feature preprocessing unit, which could be merely a MLP that operates on the node features and the adjacency matrix.
The second layer~\cref{eq:diff} is the implicit graph diffusion layer induced by the fixed-point equation~\cref{eq:equi} with source constraints on node features.
The last layer~\cref{eq:model-out} is the output layer.

\begin{remark}[DIGNN-$\Delta_\Phi$ avoids OST and OSI]
    Let $\Delta_\Phi$ be the graph neural Laplacian defined in \cref{sec:model:lap}. Denote the DIGNN architecture with $\Delta \equiv \Delta_\Phi$ by DIGNN-$\Delta_\Phi$. Given that $\Delta_\Phi$ depends on the node features $\mX$, the solution $\mZ^\star$ of \cref{eq:diff} would depend on $\mX$, which avoids the conditions in \cref{thrm:corol1}. 
    On the other hand, we have imposed source constraint on $\widetilde{\mX}$, i.e., $\gI \neq \emptyset$, which avoids the conditions in \cref{thrm:lem5}. 
\end{remark}

\subsection{Convergence and generalization of DIGNNs}\label{sec:model-bound}
The subsequent corollary provides the tractable well-posedness conditions of our implicit GNN layer and outlines an iterative algorithm designed to obtain the equilibrium, complete with an assurance of convergence.

\begin{corollary}[Tractable Well-Posedness Condition and Convergence Rate]\label[corollary]{thrm:corol3}
    Given an undirected graph $\gG = (\gV, \gE, \mA)$ with node embeddings $\widetilde{\mX} := (\widetilde{\vx}_1, \dots, \widetilde{\vx}_n)^\top \in \R^{n \times d}$, Let $\lambda_{\text{max}}$ be the largest eigenvalues of the matrix $\Delta$. Suppose that \cref{assump:graph} holds, then, the fixed-point equilibrium equation $\mZ = \widetilde{\mX} - \frac{1}{\mu}\Delta\mZ$ has a unique solution if $\mu > \lambda_{\text{max}}$. The solution can be obtained by iterating:
    \begin{equation}
        \mZ^{(t+1)} = \widetilde{\mX} - \frac{1}{\mu}\Delta\mZ^{(t)}, \text{ with } \mZ^{(0)} = \mathbf{0}, t = 0, 1, \dots \label{eq:iter}
    \end{equation}
    Therefore, $\mZ^\star = \lim_{t \rightarrow \infty}\mZ^{(t)}$. Suppose that $\|\mZ^\star\| \leq c_Z \in \R_+^*$, then $\forall t \geq 1$, $\|\mZ^{(t)} - \mZ^\star\| \leq c_Z\left(\lambda_{\text{max}} / \mu\right)^t$. Specifically,
    \begin{enumerate}\setlength{\itemsep}{-2pt}
        \item If $\Delta$ is the random walk Laplacian, i.e., $\Delta \equiv \Delta^{\text{(rw)}}$, $\mu > 2$ and $\|\mZ^{(t)} - \mZ^\star\| \leq c_Z\left(2 / \mu\right)^t$;

        \item If $\Delta$ is learnable and $\Delta \equiv \Delta_\Phi$, suppose further that \cref{assump:embedding,assump:param} hold, then we have $\mu > 2c_\phi^2c_\chi c_X\cosh(c_\chi c_X)$ and $\|\mZ^{(t)} - \mZ^\star\| \leq c_Z(2c_\phi^2c_\chi c_X\cosh(c_\chi c_X) / \mu)^t$.
    \end{enumerate}
\end{corollary}

\begin{remark}
    \cref{assump:graph} is widely used in graph learning problems.
    \cref{thrm:corol3} shows that the selection of the hyperparameter $\mu$ is linked to the upper bound $c_X, c_\chi, c_\phi$ of the norm of $\widetilde{\mX}$ and $\Theta_\chi, \Theta_\phi$ when we apply $\Delta \equiv \Delta_\Phi$. 
    To ensure \cref{assump:embedding,assump:param} hold and maintain a minimal $c_X, c_\chi, c_\phi$, we can perform \textit{feature normalization} on $\widetilde{\mX}$ and \textit{weight decay} on the parameters during model training.
\end{remark}

\textbf{Transductive learning on graphs.} 
Given an undirected graph $\gG = (\gV, \gE, \mA)$ with node features and labels $\{(\vx_i, y_i)\}_{i=1}^{m+u}$ where $m+u = n$. 
Without loss of generality, let $\{y_i\}_{i=1}^m$ be the selected node labels, we aim to predict the labels of all nodes by a learner (model) trained on the graph $\gG$ with $\{\vx_i\}_{i=1}^{m+u} \cup \{y_i\}_{i=1}^m$. 
For any $f \in \gH$, the training and test error is defined as: $\widehat{L}_m(f) := \frac{1}{m}\sum_{i=1}^m\ell(f(\vx_i), y_i), L_u(f) := \frac{1}{u}\sum_{i=m+1}^{m+u}\ell(f(\vx_i), y_i)$, respectively, where $\ell: \gH \times \gX \times \gY \mapsto \R_+$ is the loss function. 
We are interested in the transductive generalization gap of $f$ which is defined as $|\widehat{L}_m(f) - L_u(f)|$.

\begin{assumption}\label{assump:loss}
    Loss function $\ell$ is $\rho_\ell$-Lipschitz continuous.
\end{assumption}

\begin{assumption}\label{assump:output}
    The output layer~\cref{eq:model-out} of DIGNN, i.e., $h_{\Theta^{(2)}}$, is $\rho_h$-Lipschitz continuous.
\end{assumption}

\begin{remark}
\cref{assump:loss} is satisfied by some commonly used loss functions, such as mean square error and cross-entropy loss.
\cref{assump:output} is also relatively modest, as the output layer is typically composed of a linear function and a softmax function, which exhibits Lipschitz continuity.
\end{remark}

\begin{theorem}[Transductive Generalization Bounds for DIGNN-$\Delta_\Phi$]\label{thrm:dignn-generalization}
    Let $\Delta_\Phi$ be defined by functions $\chi, \phi, \varphi$ as given in \cref{eq:eq12,eq:eq13,eq:eq14} with learnable parameters $\Theta_\chi \cup \Theta_\phi \cup \Theta_\varphi =: \Phi \in \Omega$, $\lambda_\Phi$ be the largest eigenvalue of $\Delta_\Phi$.
    Let \cref{eq:model-in,eq:model-out,eq:iter} be the input, the output, and the implicit graph diffusion layers of DIGNN-$\Delta_\Phi$ respectively.
    Suppose that \cref{assump:loss,assump:output} hold and the node embeddings $\widetilde{\mX}$ of the first layer (\cref{eq:model-in}) of DIGNN-$\Delta_\Phi$ satisfy \cref{assump:embedding}. Then, for any integer $T \geq 1$, any $\delta > 0$, with probability of at least $1-\delta$ over the choice of the training set from $\gG$, for all DIGNN-$\Delta_\Phi$ model $f \in \gH$,
    \begin{equation}
        L_u(f) \leq \widehat{L}_m(f) + \rho_\ell\rho_h\frac{c_X}{\sqrt{n}}\sum_{t=0}^{T-1}\left(\frac{\lambda_\Phi}{\mu}\right)^{t} + c_0c_1\sqrt{\min(m, u)} + \sqrt{\frac{c_1c_2}{2}\log\frac{1}{\delta}}, \label{eq:dignn-gen1}
    \end{equation}
    where $c_0 := \sqrt{32\ln(4e) /3} < 5.05, c_1 := (\frac{1}{m} + \frac{1}{u})$, and $c_2 := \frac{m+u}{(m+u-1/2)(1-1/2(\max(m, u)))}$.
    Suppose further that \cref{assump:graph,assump:param} hold, we have
    \begin{equation}
        L_u(f) \leq \widehat{L}_m(f) +  \rho_\ell\rho_h\frac{c_X}{\sqrt{n}}\sum_{t=0}^{T-1}\left(\frac{2c_\phi^2c_\chi c_X\cosh(c_\chi c_X)}{\mu}\right)^{t} + c_0c_1\sqrt{\min(m, u)} + \sqrt{\frac{c_1c_2}{2}\log\frac{1}{\delta}}. \label{eq:dignn-gen2}
    \end{equation}
\end{theorem}

\begin{remark}[Connections between Convergence, Generalization, and Feature Smoothing]\label{remk:connection}
    As discussed in \cref{sec:analysis-obj}, the hyperparameter $\mu$ controls the trade-off between smoothing and constraints on source nodes.
    As $\mu$ decreases, greater emphasis is imposed on feature smoothing, and vice versa.
    On the other hand, \cref{thrm:corol3} and \cref{thrm:dignn-generalization} demonstrate that both the graph diffusion convergence rates and generalization bounds of DIGNN-$\Delta_\Phi$ depend on the largest eigenvalue $\lambda_\Phi$ of $\Delta_\Phi$ and $\mu$.
    DIGNN-$\Delta_\Phi$ can achieve faster convergence rates and better generalization bounds as the ratio $\frac{\lambda_\Phi}{\mu}$ decreases.
    It reveals a significant link between convergence and generalization properties, indicating that superior generalization ability for DIGNN-$\Delta_\Phi$ requires fast convergence for the implicit graph diffusion.
    Thus, an appropriately chosen $\mu$ could not only provide a suitable balance between feature smoothing and source constraints but also ensure reliable convergence and generalization properties.
\end{remark}

Furthermore, as $T$ goes to infinity, the graph neural diffusion would converge to the equilibrium and the generalization bound converges to the form of \cref{eq:dignn-gen3}.
We summarize the convergence and generalization analysis in \cref{thrm:dignn-main}:

\begin{corollary}[Convergence and Generalization of DIGNN-$\Delta_\Phi$]\label{thrm:dignn-main}
    Under the same settings of \cref{thrm:dignn-generalization}, denote by $\mZ^\star$ the optimal solution of $\mZ = \widetilde{\mX} - \frac{1}{\mu}\Delta_\Phi\mZ$. 
    Suppose that \cref{assump:graph,assump:embedding,assump:param,assump:loss,assump:output} hold and $\|\mZ^\star\| \leq c_Z$. 
    Let $\mu > 2c_\phi^2c_\chi c_X\cosh(c_\chi c_X)$. For any $c_Z > \epsilon > 0, \delta > 0$, let $T = \frac{\log c_Z - \log\epsilon}{-\log\left(2c_\phi^2c_\chi c_X\cosh(c_\chi c_X) / \mu\right)}$, then $\|\mZ^{(T)} - \mZ^\star\| \leq \epsilon$ and \cref{eq:dignn-gen2} holds with probability at least $1-\delta$ over the choice of the training set from $\gG$. Moreover, for $T \rightarrow \infty$, we have $\lim_{T \rightarrow \infty}\mZ^{(T)} = \mZ^\star$ and with probability at least $1-\delta$, for all DIGNN-$\Delta_\Phi$ model $f \in \gH$,
    \begin{equation}
        L_u(f) \leq \widehat{L}_m(f) +  \frac{\rho_\ell\rho_hc_X}{\sqrt{n}(1 - 2c_\phi^2c_\chi c_X\cosh(c_\chi c_X) / \mu)} + c_0c_1\sqrt{\min(m, u)} + \sqrt{\frac{c_1c_2}{2}\log\frac{1}{\delta}}. \label{eq:dignn-gen3}
    \end{equation}
\end{corollary}

\subsection{Training of DIGNNs}
For the \textit{forward evaluation}, we can simply iterate \cref{eq:iter} to obtain the equilibrium in terms of \cref{thrm:corol3}.
For the \textit{backward pass}, we can use implicit differentiation~\citep{ImplicitBook-2002,DEQ-NIPS2019} to compute the gradients of trainable parameters by directly differentiating through the equilibrium.
We refer to \cref{app:train-dignn} for a more detailed illustration of the forward and backward pass for DIGNNs.

We discuss the computational complexity and running time of DIGNNs in \cref{app:complex}.
\section{Experimental Results}\label{sec:exp}
In this section, we conduct experiments to evaluate the effectiveness of our models against implicit and explicit GNNs on benchmark datasets for node and graph classification tasks. 
See \cref{app:exp} for the details of data statistics and experimental setup. 
We adopt the random walk Laplacian $\Delta^{\text{(rw)}}$ (discussed in \cref{app:lap}) and the graph neural Laplacian $\Delta_\Phi$ for DIGNN (termed as DIGNN-$\Delta^{\text{(rw)}}$ and DIGNN-$\Delta_\Phi$ respectively).
Our models are implemented based on the PyTorch Geometric library~\citep{pyg-aXiv2019}. 

\subsection{Semi-supervised node classification}
\textbf{Datasets.} 
We conduct node classification experiments on five heterophilic datasets: Chameleon, Squirrel~\citep{NodeHeterData-JCN2021}, Penn94, Cornell5, and Amherst41~\citep{LINKX-NIPS2021}. 
For Chameleon, Squirrel, we use standard train/validation/test splits as in \citep{Geom-GCN-ICLR2020}. 
For Penn94, Cornell15, and Amherst41, we follow the same data splits as in \citep{LINKX-NIPS2021}.
We also conduct experiments on three commonly used homophilic datasets: Cora, CiteSeer, PubMed~\citep{NodeHomoData-AIM2008} and PPI dataset~\citep{NodePPIData-Bio2017} for multi-label multi-graph inductive learning, which are presented in \cref{tab:tab3,tab:tab4} in \cref{app:exp-homo}.

\textbf{Baselines.} 
We compare our models with several representative explicit GNNs such as GCN~\citep{GCN-ICLR2017}, GAT~\citep{GAT-ICLR2018}, GCNII~\citep{GCNII-ICML2020}, H2GCN~\citep{H2GCN-NIPS2020}, APPNP~\citep{APPNP-ICLR2019}, LINKX~\citep{LINKX-NIPS2021}, and implicit GNNs, including IGNN~\citep{IGNN-NIPS2020}, GRAND-l~\citep{GRAND-ICML2021}, MGNNI~\citep{MGNNI-NIPS2022}, GIND~\citep{GIND-ICML2022}.
The results of all baselines except for GRAND-l on Chameleon, Squirrel are borrowed from \cite{MGNNI-NIPS2022,GIND-ICML2022,LINKX-NIPS2021}, while the results for GRAND-l were obtained by us using their officially published source code. 
For Penn94, Cornell5, and Amherst41, all baselines were run by us to obtain their results.

\begin{table}[tp]
    \centering
    \caption{Results on heterophilic graph datasets: mean accuracy (\%) (standard deviation (stdev)) over 10 random data splits. Best results outlined in bold. OOT denotes out of time.}\label{tab:tab2}
    \resizebox{\linewidth}{!}%
    {
    \begin{tabular}{lccccc}
    \toprule
    & Chameleon & Squirrel & Penn94 &   Cornell5 & Amherst41  \\
    \midrule
    \#Nodes & 2,227 & 5,201 & 41,554 & 18,660 & 2,235 \\
    \#Edges & 31,421 & 198,493 & 1,362,229 &  790,777 & 90,954 \\
    \#Clases & 5 & 5 & 2 & 2 & 2 \\
    \midrule
     GCN & $42.34$ ($\gray{2.77}$) & $29.00$ ($\gray{1.10}$) &   $82.47$ ($\gray{0.27}$)  & $80.15$ ($\gray{0.37}$) & $81.41$ ($\gray{1.70}$) \\
     GAT & $46.03$ ($\gray{2.51}$) & $30.51$ ($\gray{1.28}$) &  $81.53$ ($\gray{0.55}$) & $78.96$ ($\gray{1.57}$) &  $79.33$ ($\gray{2.09}$) \\
     GCNII   & $48.59$ ($\gray{1.88}$) & $32.20$ ($\gray{ 1.06}$) & $82.92$ ($\gray{0.59}$) & $78.85$ ($\gray{0.78}$) & $76.02$ ($\gray{1.38}$) \\  
     H2GCN & $60.30$ ($\gray{ 2.31}$) & $40.75$ ($\gray{1.44}$) & $81.31$ ($\gray{0.60}$) & $78.46$ ($\gray{0.75}$) & $79.64$ ($\gray{1.63}$) \\
    APPNP & $43.85$ ($\gray{2.43}$) & $30.67$ ($\gray{1.06}$) & $74.79$ ($\gray{0.43}$) & $73.23$ ($\gray{1.08}$) & $68.34$ ($\gray{2.92}$) \\
    LINKX & $68.42$ ($\gray{1.38}$) & $61.81$ ($\gray{1.80}$) & $84.71$ ($\gray{0.52}$) & $83.46$ ($\gray{0.61}$) & $81.73$ ($\gray{1.94}$) \\
    \midrule
    IGNN & $41.38$ ($\gray{2.53}$) & $24.99$ ($\gray{2.11}$) & - & - &  -  \\
    GRAND-l* & $42.39$ ($\gray{4.61}$) & $34.57$ ($\gray{0.89}$) & $81.77$ ($\gray{0.39}$) & $81.22$ ($\gray{0.81}$) & $80.07$ ($\gray{1.99}$) \\
    MGNNI & $63.93$ ($\gray{2.21}$) & $54.50$ ($\gray{ 2.10}$) & OOT & $78.11$ ($\gray{0.67}$) & $75.59$ ($\gray{1.65}$) \\
    GIND & $66.82$ ($\gray{2.37}$) & $56.71$ ($\gray{2.07}$) & $76.29$ ($\gray{0.78}$) & $73.91$ ($\gray{0.92}$) & $72.24$ ($\gray{2.20}$) \\
    \midrule
    DIGNN-$\Delta^{\text{(rw)}}$ & $79.21$ ($\gray{1.22}$) & $73.76$ ($\gray{2.19}$) & $83.17$ ($\gray{0.33}$) & $82.27$ ($\gray{0.56}$) &  $80.78$ ($\gray{2.17}$) \\
    DIGNN-$\Delta_\Phi$ & $\bf{79.89}$ ($\gray{1.51}$) & $\bf{74.96}$ ($\gray{1.77}$) & $\bf{86.19}$ ($\gray{0.18}$) & $\bf{84.44}$ ($\gray{0.59}$) & $\bf{83.34}$ ($\gray{1.21}$) \\
    \bottomrule
    \end{tabular}
    }
\end{table}

\textbf{Results.} The results in \cref{tab:tab2} illustrate that DIGNN-$\Delta_\Phi$ substantially surpasses both explicit and implicit baselines on all heterophilic datasets.
In particular, our model DIGNN-$\Delta_\Phi$ considerably improves the accuracy of LINKX on Chameleon from $68.42\%$ to $79.89\%$ and Squirrel from $61.81\%$ to $74.96\%$.
In contrast, the best implicit GNN baseline, GIND, only managed to achieve $66.82\%$ and $56.71\%$ respectively.
Additionally, the results for Penn94, Cornell5, and Amherst41 state that DIGNN-$\Delta_\Phi$ consistently surpasses all explicit and implicit baselines.
Moreover, APPNP, GCNII, and implicit baselines, which can consider a larger range of neighbors, do not outperform GCN, GAT, LINKX.
It suggests that for these three datasets, long-range neighborhood information may not be as beneficial as they are for Chameleon and Squirrel.
Instead, the metrics of vertex and edge spaces may play a more significant role.
The conclusion can be reinforced by observing that DIGNN-$\Delta_\Phi$ surpasses DIGNN-$\Delta^{\text{(rw)}}$, which adopts fixed vertex and edge Hilbert spaces and cannot learn the graph metrics.

\begin{table}[tp]
    \centering
    \caption{Results on graph datasets: accuracy (\%) (stdev) over 10 folds. Best results outlined in bold.}\label{tab:tab5}
    \resizebox{\linewidth}{!}%
    {
    \begin{tabular}{lcccccc}
        \toprule
        & MUTAG & PTC & PROTEINS & NCI1 & IMDB-B & IMDB-M \\
        \#Graphs & 188 &  344  & 1113 & 4110 & 1000 & 1500 \\
        Avg \#Nodes & 17.9 & 25.5  & 39.1 & 29.8 & 19.8 & 13.0 \\
        \#Classes & 2  &  2  & 2 & 2 & 2 & 3 \\
        \midrule
        GCN & $85.6$ ($\gray{5.8}$) & $64.2$ ($\gray{4.3}$) & $76.0$ ($\gray{3.2}$) & $80.2$ ($\gray{2.0}$) & - & - \\
        GIN & $89.0$ ($\gray{6.0}$) & $63.7$ ($\gray{8.2}$) & $75.9$ ($\gray{3.8}$) & $\bf{82.7}$ ($\gray{1.6}$) & $75.1$ ($\gray{5.1}$) & $52.3$ ($\gray{2.8}$) \\
        DGCNN & $85.8$ & $58.6$ & $75.5$ & $74.4$ & $70.0$ & $47.8$ \\
        FDGNN & $88.5$ ($\gray{3.8}$) & $63.4$ ($\gray{5.4}$) & $76.8$ ($\gray{2.9}$) & $77.8$ ($\gray{1.6}$) & $72.4$ ($\gray{3.6}$) & $50.0$ ($\gray{1.3}$) \\
        \midrule
        IGNN & $89.3$ ($\gray{6.7}$) & $70.1$ ($\gray{5.6}$) & $77.7$ ($\gray{3.4}$) & $80.5$ ($\gray{1.9}$) & - & - \\
        EIGNN & $88.9$ ($\gray{1.1}$) & $69.8$ ($\gray{5.3}$) & $75.9$ ($\gray{6.4}$) & $77.5$ ($\gray{2.2}$) & $72.3$ ($\gray{4.3}$) & $52.1$ ($\gray{2.9}$) \\
        CGS & $89.4$ ($\gray{5.6}$) & $64.7$ ($\gray{6.4}$) & $76.3$ ($\gray{6.3}$) & $77.2$ ($\gray{2.0}$) & $73.1$ ($\gray{3.3}$) & $51.1$ ($\gray{2.2}$) \\
        MGNNI & $91.9$ ($\gray{5.5}$) & $72.1$ ($\gray{2.8}$) & $79.2$ ($\gray{2.9}$) & $78.9$ ($\gray{2.1}$) & $75.8$ ($\gray{3.4}$) & $53.5$ ($\gray{2.8}$) \\
        GIND & $89.3$ ($\gray{7.4}$) & $66.9$ ($\gray{6.6}$) & $77.2$ ($\gray{2.9}$) & $78.8$ ($\gray{1.7}$) & - & - \\
        \midrule
        DIGNN-$\Delta^{\text{(rw)}}$ & $90.8$ ($\gray{7.4}$) & $74.1$ ($\gray{5.3}$) & $79.9$ ($\gray{2.9}$) & $77.7$ ($\gray{1.4}$)  & $77.7$ ($\gray{3.2}$) & $53.9$ ($\gray{3.7}$)  \\
        DIGNN-$\Delta_\Phi$ & $\bf{94.6}$ ($\gray{4.9}$) & $\bf{76.5}$ ($\gray{3.7}$)  & $\bf{81.3}$ ($\gray{3.2}$) & $77.9$ ($\gray{2.0}$ & $\bf{78.4}$ ($\gray{3.6}$) & $\bf{54.2}$ ($\gray{2.9}$) \\
        \bottomrule
    \end{tabular}
    }
\end{table}

\subsection{Graph classification}
\textbf{Datasets.} We use four bioinformatics datasets~\citep{GraphBioData-KDD2015}, MUTAG, PTC, PROTEINS, NCI1, and two social network datasets~\citep{GraphBioData-KDD2015}, IMDB-BINARY and IMDB-MULTI. We follow the standard train/validation/test splits as in \citep{GIN-ICLR2019}.

\textbf{Baselines.} We compare our models with explicit GNN baselines, i.e., GCN, GIN~\citep{GIN-ICLR2019}, DGCNN~\citep{DGCNN-AAAI2018}, FDGNN~\citep{FDGNN-AAAI2020}, and implicit GNN baselines, i.e., IGNN, CGS~\citep{CGS-ICLR2022}, MGNNI, GIND. The results for all baselines are borrowed from \citep{MGNNI-NIPS2022} and \citep{GIND-ICML2022}.

\textbf{Results.} As shown in \cref{tab:tab5}, DIGNN-$\Delta_\Phi$ substantially outperforms all baselines on all graph classification datasets except for NCI1.
For NCI1, DIGNNs ($\Delta^{\text{(rw)}}$ and $\Delta_\Phi$) are comparable to the baselines. 
The implicit baselines, MGNNI and GIND, are better than explicit baselines on most datasets. 
It suggests that long-range neighborhood information could be helpful for graph classification tasks.
Moreover, DIGNN-$\Delta_\Phi$ outperforms $\Delta^{\text{(rw)}}$ and implicit baselines on all datasets. 
The results demonstrate that the underlying metrics of vertex and edge spaces is also important for graph classification when generalizing to unseen testing graphs. 
These observations again confirm the capability of DIGNN-$\Delta_\Phi$ in effectively learning from graphs.

\begin{figure}
    \centering
    \includegraphics[width=\linewidth]{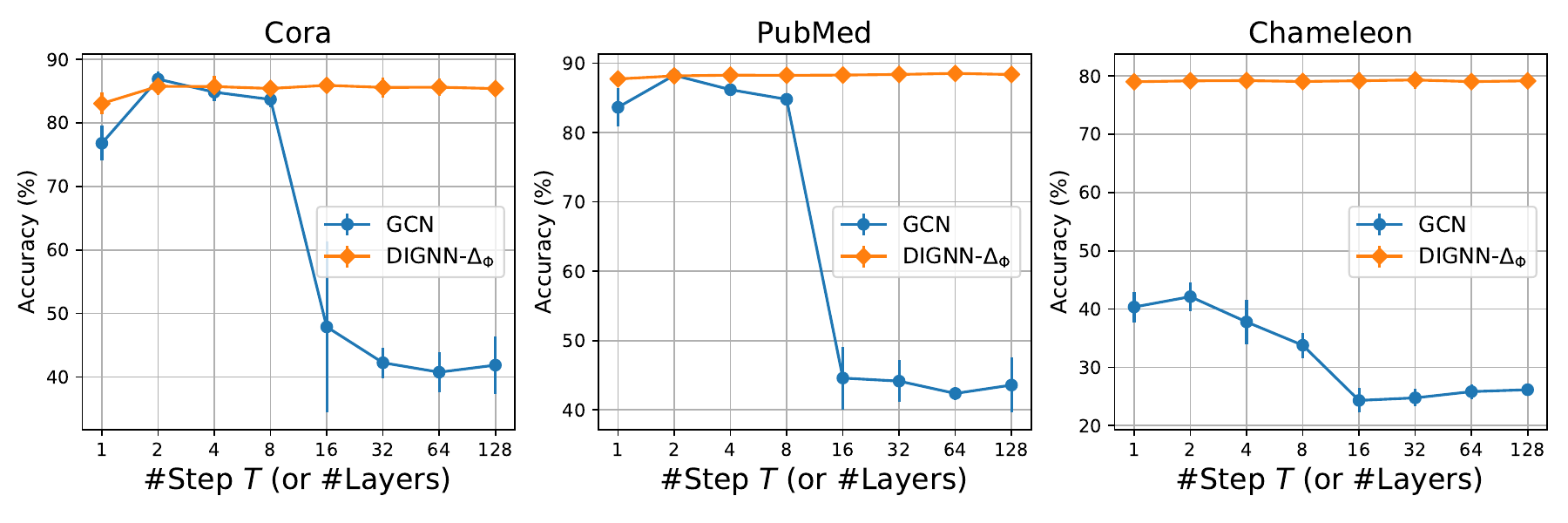}
    \caption{Over-smoothing analysis w/\ \#step $T$ (or \#layers).}\label{fig:os-analysis}
\end{figure}

\subsection{Empirical validation of theoretical results}
\textbf{Over-smoothing.} 
We carry out node classification experiments on Cora, PubMed, and Chameleon datasets to show our model would not suffer from over-smoothing and performs well with large iteration steps $T$. 
\cref{fig:os-analysis} illustrates that as the number of steps (or layers for GCN) increases, our model maintains good performance whilst GCN degrades catastrophically. 
It shows that DIGNN-$\Delta_\Phi$ can avoid over-smoothing for Cora, PubMed, and Chameleon datasets.

\begin{figure}
    \centering
    \includegraphics[width=\linewidth]{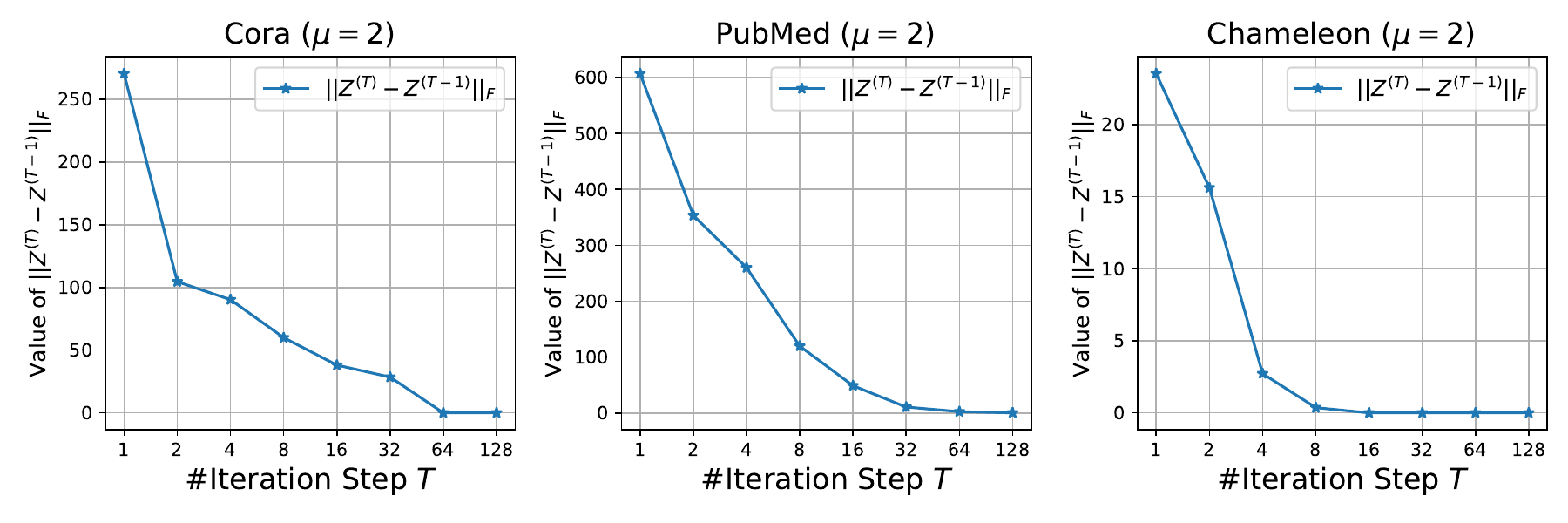}
    \caption{Convergence analysis w/\ \#step $T$.}\label{fig:T-pubmed}
\end{figure}

\textbf{Convergence analysis.}
We conduct experiments on Cora, PubMed, and Chameleon datasets to investigate the convergence properties of the implicit layers for DIGNN-$\Delta_\Phi$ concerning different iteration numbers $T$.
We use $\mu = 2$ and other hyperparameters from \cref{tab:tab7}. 
The results of \cref{fig:T-pubmed} state as $T$ increases, the value of $\|\mZ^{(T)} - \mZ^{(T-1)}\|_F$ on three datasets quickly decreases and approaches to $0$.
It validates the convergence results given in \cref{thrm:corol3}.

\begin{figure}
    \centering
    \includegraphics[width=\linewidth]{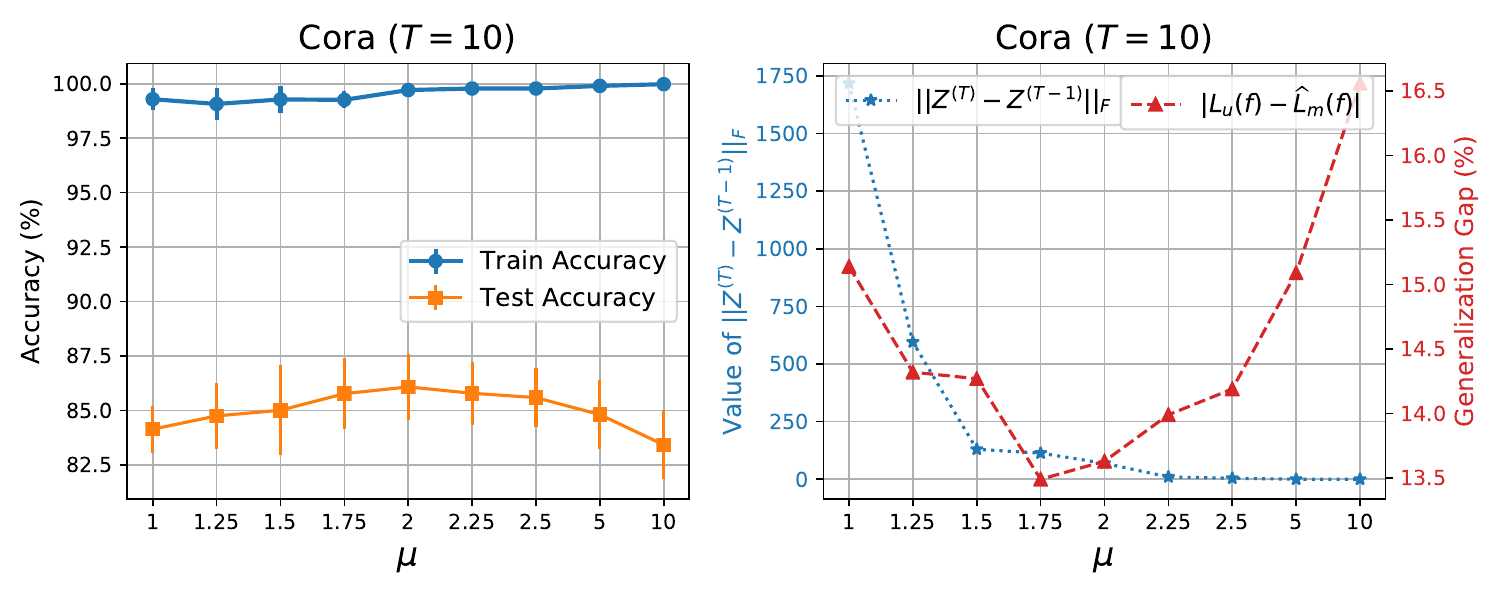}
    \includegraphics[width=\linewidth]{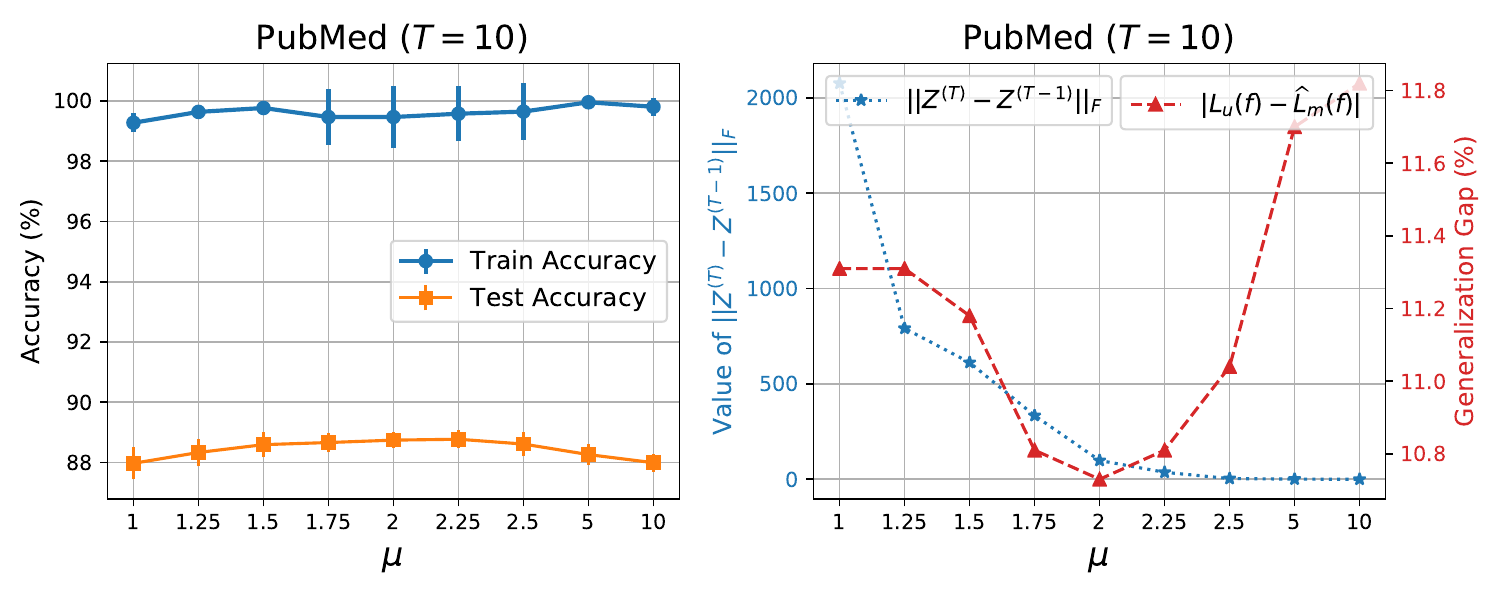}
    \caption{The impact of $\mu$ for Cora and PubMed datasets.}\label{fig:mu_analysis}
\end{figure}

\textbf{The impact of hyperparameter $\mu$.}
We study the impact of $\mu$ on the performance of DIGNN-$\Delta_\Phi$ for Cora and PubMed datasets.
We use $T=10$ and other hyperparameters following \cref{tab:tab7} and fix the spectral norm of $\Theta_\chi, \Theta_\phi$ at $1$ by applying \textit{spectral normalization} on them.
The results of PubMed in \cref{fig:mu_analysis} indicate that as $\mu$ increases from $1$ to $2$, both the term $\|\mZ^{(T)} - \mZ^{(T-1)}\|_F$ and the generalization gap $|L_u(f) - \widehat{L}_m(f)|$ decreases.
Similar observation to the results of Cora.
This indicates that, within a certain range, as $\mu$ increases ($\frac{\lambda_\Phi}{\mu}$ will decrease as $\|\Theta_\chi\|, \|\Theta_\phi\|$ are fixed), DIGNN-$\Delta_\Phi$ will have better convergence and generalization properties.
It also validates the discussion in \cref{remk:connection} that fast convergence results in better generalization.

When $\mu > 2$, $\|\mZ^{(T)} - \mZ^{(T-1)}\|_F$ approaches to $0$ but the generalization gap increases.
It is reasonable as for $\mu > 2$, DIGNN-$\Delta_\Phi$ already converges well and its performance becomes more influenced by other factors.
As $\mu$ increases, \cref{eq:Dirichlet} would impose more emphasis on ensuring the input feature constraints and less on feature smoothing.
However, overly emphasizing the constraints, e.g., $\mu \geq 5$, could lead to overfitting in the input and output layers described in \cref{eq:model-in} in \cref{eq:model-out} respectively, which are assumed to exhibit constant Lipschitness in our theoretical analysis.

These results support our claim in \cref{remk:connection} that $\mu$ not only controls the trade-off between feature smoothing and input constraints in \cref{eq:Dirichlet}, but also influences the convergence and generalization properties for the model.

\section{Conclusion}\label{sec:con}
This paper introduces a geometric framework to equip implicit GNNs with the ability to learn the metrics of vertex and edge spaces while avoiding over-smoothing and ensuring reliable convergence and generalization properties.
We further propose a new implicit GNN model based on our framework and derive new convergence and generalization results.
Our model empirically works well on benchmark datasets for both node and graph classification tasks.

\bibliography{reference}
\bibliographystyle{apalike}


\clearpage
\appendix

\appendixtitle{Appendix}

\etocdepthtag.toc{mtappendix}
\etocsettagdepth{mtchapter}{none}
\etocsettagdepth{mtappendix}{subsection}

\tableofcontents

\clearpage

\section{Related Work and Discussion}\label{sec:app:related}
In this section, we discuss the related work and highlight our contributions against existing approaches. 
Our work generally pertains to several areas, including implicit GNNs, graph neural diffusions, Over-smoothing analysis and generalization bounds for GNNs, and graph-based semi-supervised learning.

\textbf{Implicit GNNs.} 
Instead of stacking a series of propagation layers hierarchically, implicit GNNs define their outputs as solutions to fixed-point equations, which are equivalent to running infinite-depth GNNs. 
As a result, implicit GNNs are able to capture long-range dependencies and benefit from the global receptive field. 
Moreover, by using implicit differentiation~\citep{ImplicitBook-2002} for the backward pass gradient computation, implicit GNNs require only constant training memory~\citep{DEQ-NIPS2019,IDL-aXiv2019}. 
A series of implicit GNN architectures, e.g., \citep{GNN-TNN2009,IGNN-NIPS2020,EIGNN-NIPS2021,CGS-ICLR2022,GIND-ICML2022,MGNNI-NIPS2022}, have been proposed. 
Probably one of the earliest implicit GNN models, proposed by \citep{GNN-TNN2009}, is based on a constrained information diffusion mechanism and it guarantees the existence of a unique stable equilibrium.
The implicit layers of IGNN~\citep{IGNN-NIPS2020} and EIGNN~\citep{EIGNN-NIPS2021} are developed based on the aggregation scheme of GCN~\citep{GCN-ICLR2017},  which corresponds to an isotropic linear diffusion operator that treats all neighbors equally and is independent of the neighbor features~\citep{DGC-NIPS2021,GIND-ICML2022}. 
CGS~\citep{CGS-ICLR2022} adopts an input-dependent learnable propagation matrix in their implicit layer which ensures guaranteed convergence and uniqueness for the fixed-point solution. 
However, \citep{GNNExpress-ICLR2020} shows that isotropic linear diffusion operators would lead to the over-smoothing issue. 
Therefore, GIND~\citep{GIND-ICML2022} designs a non-linear diffusion operator with anisotropic properties whose equilibrium state corresponds to a convex objective and the anisotropic property may mitigate over-smoothing. 
Additionally,  to incorporate multiscale information on graphs, MGNNI~\citep{MGNNI-NIPS2022} expands the effective range of EIGNN for capturing long-range dependencies to multiscale graph structures.

\textbf{Graph neural diffusions.} 
GNNs can be interpreted as diffusion processes on graphs. 
\citep{DiffGNN-NIPS2016,DiffGNN-NIPS2019} design their graph convolutions based on diffusion operators and \citep{LanczosNet-ICLR2019} introduces a polynomial graph filter to exploit multi-scale graph information which is closely related to diffusion maps. 
On the other hand, neural differential equations~\citep{NDE-NIPS2018,NDE-arXiv2022} have also been applied to design graph neural diffusion processes. 
Continuous depth GNNs, such as CGNN~\citep{CGNN-ICML2020}, GRAND~\citep{GRAND-ICML2021}, PDE-GCN~\citep{PDE-GCN-NIPS2021}, have been developed based on the discretization of diffusion partial differential equations (PDEs) on graphs. 
\citep{GRAND++-ICLR2022} introduces a source term into the diffusion process of GRAND to mitigate the over-smoothing issue. 
\citep{DGC-NIPS2021} shows the equivalence between the propagation scheme of GCN and the numerical discretization of an isotropic diffusion process.
Specific PDEs could be helpful to improve the capability of GNNs in capturing the graph metrics or mitigating the over-smoothing issue. 
For example, \citep{Beltrami-NIPS2021} develops graph diffusion equations based on the discretized Beltrami flow which facilitates feature learning, positional encoding, and graph rewiring within the diffusion process. 
\cite{OS-NIPS2022} employs cellular sheaf theory to develop neural diffusion processes capable of handling heterophilic graphs and addressing the over-smoothing issue.

While the above implicit GNN and graph neural diffusion models have demonstrated their effectiveness in many applications, they have mainly concentrated on learning the diffusion strength without explicitly engaging in learning the metrics of vertex and edge spaces. 
The underlying graph metric has been shown to have a profound connection with the performance of GNNs in heterophilic settings and their over-smoothing behaviours~\citep{Sheaf-NIPS2022}. 
These methods may lack of sufficient adaptability in learning the graph metrics, which could potentially hinder their performance in more broad graph learning problems.
On the contrary, we use three parametrized positive real-valued functions to explicitly learn the Hilbert spaces for vertices, the Hilbert space for edges, and the graph diffusion strength on each edge, respectively. 
Therefore, our model could have more flexibility in modeling the graph metrics.

\textbf{Over-Smoothing Analysis of GNNs.}
Over-smoothing issues have been widely studied in deep GNN architectures.
A series of works, e.g., \citet{OS-AAAI2018,OS-CIKM2021,OS-NIPS2022,OS/iclr/ZhaoA20,DGC-NIPS2021,OS/iclr/WuCWJ23}, have demonstrated that as the number of explicit GNN layers goes to infinity, the representations will converge to the same value across all nodes, which we call over-smoothing issue during training (OST).
In this work, we characterize the conditions for implicit GNN layers, i.e., if the graph parameterizations are not defined as functions of the input node features, that will lead to the same OST issue.
Additionally, \citet{GRAND++-ICLR2022} has presented a random walk interpretation of GRAND, showing that the node features will become the same across all nodes under the GRAND-l dynamics.
Another type of issue we call over-smoothing during inference (OSI).
Our work shows that the same OSI issue could present in implicit GNN architectures for the task without node label constraints during the inference process, such as graph classification.

\textbf{Generalization Bounds of GNNs.} 
Numerous studies have established generalization bounds for explicit GNNs. 
They can categorized into the bounds derived based on 
Vapnik–Chervonenkis dimension~\citep{gener/nn/ScarselliTH18}, Rademacher complexity~\citep{gener/icml/GargJJ20,gener/nips/OonoS20,gener/nips/EsserVG21,gener/icml/tang23}, PAC-bayesian~\citep{gener/iclr/LiaoUZ21,gener/aistats/JuLSZ23}, algorithmic stability~\citep{gener/kdd/VermaZ19,gener/ijon/ZhouW21,gener/nips/CongRM21}, neural tangent kernel~\citep{gener/nips/DuHSPWX19}, and many others~\citep{gener/icml/YehudaiFMCM21,gener/icml/BaranwalFJ21,gener/icml/LiWLCX22,gener/nips/MaskeyLLK22}. 
However, the generalization analysis for implicit GNNs is relatively overlooked and the emphasis in existing implicit GNN literature is only on ensuring convergence properties.
In this work, we derive new convergence and generalization results for our implicit GNN model.
Specifically, we demonstrate that with an appropriately chosen hyperparameter greater than the largest eigenvalue of the parameterized graph Laplacian, our proposed implicit graph diffusion layer is guaranteed to converge to the equilibrium.
We also derive the transductive generalization bounds for our model based on the transductive Rachamecher complexity.
Our theoretical results further reveal a significant link between convergence and generalization properties for implicit GNNs, indicating that superior transductive generalization ability requires fast convergence for implicit graph diffusion.

\textbf{Graph-based semi-supervised learning.} 
Graph-based semi-supervised learning algorithms usually assume that the labels of neighboring nodes in a graph are the same or consistent. 
Over the past few decades, numerous methods have been proposed.
They generally intend to minimize a smoothness regularizer with label constraints at some vertices to ensure label consistency among neighboring nodes.
Such a smoothness regularizer can be various, including graph Laplacian regularization~\citep{SSL-ICML2003,SSL-COLT2003,SSL-NIPS2004,SSL-COLT2004,SSL-DAGM2005,SSL-NIPS2009,SSL-arXiv2019,SSL-arXiv2023}, Tikhonov regularization~\citep{SSL-COLT2004}, manifold regularizations~\citep{SSL-JMLR2006,SSL-JMLR2013}.
We refer to \citep{SSLSurvey-Year2005,SSLSurvey-ML2020} for a more comprehensive review of graph-based semi-supervised learning.
These methods are developed based on non-parameterized graph Laplacians whose vertex and edge spaces are fixed and independent of the node features.
As we characterized in \cref{thrm:corol1}, this could lead to over-smoothing during training (OST), i.e., changing the node features will not affect the output node representations during training. 
Moreover, when it comes to graph classification or graph regression problems that do not have label constraints during the inference process, \cref{thrm:lem4} shows that their output may converge to a constant function over all nodes regardless of the input features, a problem we call over-smoothing during inference (OSI).
Grounding on the theoretical justifications, our derived model can overcome these two types of over-smoothing issues.

\textbf{Limitations.} 
One of the main limitations of our work is that our theoretical analysis and methodology have focused on linear graph neural diffusion operators. 
However, the theoretical foundation for non-linear dynamic systems remains an important impediment to the entire field of machine learning. 
Nonetheless, our work has provided some valuable
insights into the interplay between implicit GNNs, differential geometry, and semi-supervised learning. 
There are still numerous unexplored insights to be discovered and we look forward to seeing further cross-fertilization between machine learning and algebraic topology in the future.

\clearpage
\section{Theorems and Proofs}
\subsection{Theorems and proofs in \cref{sec:pre}}

\subsubsection{Proof of \cref{thrm:lem1}}\label{app:thrm:lem1}
\textbf{\cref{thrm:lem1}.}\textit{
    Given an undirected graph $\gG = (\gV, \gE)$ and a function $g: \gE \mapsto \R$, the parameterized graph divergence $\divg: \gH(\gE, \phi) \mapsto \gH(\gV, \chi)$ is explicitly given by
    \begin{equation*}
        (\divg g)(i) = \frac{1}{2\chi(i)}\sum_{j=1}^n\varphi([i, j])\phi([i, j])\left(g([i, j]) - g([j, i])\right).
    \end{equation*}
}
\begin{proof}
    For any $k = 1, 2, \dots, n$, using the indicator function $f(i) = \sI_{i=k}$, where
    \begin{equation*}
        \sI_{i = k} = \begin{cases}
        1 & \text{ if } i = k, \\
        0 & \text{ if } i \neq k.
        \end{cases}
    \end{equation*}
    By the definition of the inner product on the function space on the edges, we have
    \begin{align*}
        \langle \nabla\sI_{\cdot = k}, g \rangle_\gE = {} & \frac{1}{2}\sum_{i=1}^n\sum_{j=1}^n(\nabla\sI_{\cdot = k})([i, j]) g([i, j])\phi([i, j]) \\
        = {} & \frac{1}{2}\sum_{i=1}^n\sum_{j=1}^n\varphi([i, j])(\sI_{j=k} - \sI_{i=k})g([i, j])\phi([i, j]) \\
        = {} & \frac{1}{2}\left(\sum_{i=1}^n\sum_{j=1}^n\varphi([i, j])\sI_{j=k}g([i, j])\phi([i, j]) - \sum_{i=1}^n\sum_{j=1}^n\varphi([i, j])\sI_{i=k})g([i, j])\phi([i, j])\right) \\
        = {} & \frac{1}{2}\left(\sum_{i=1}^n\varphi([i, k])g([i, k])\phi([i, k]) - \sum_{j=1}^n\varphi([k, j])g([k, j])\phi([k,j])\right) \\
        = {} & \frac{1}{2}\left(\sum_{j=1}^n\varphi([j, k])g([j, k])\phi([j, k]) - \sum_{j=1}^n\varphi([k, j])g([k, j])\phi([k, j])\right) \tag{Replace $i$ by $j$ for the first term} \\
        = {} & \frac{1}{2}\left(\sum_{j=1}^n\varphi([k, j])\phi([k, j])g([j, k]) - \sum_{j=1}^n\varphi([k, j])\phi([k, j])g([k, j])\right) \tag{$[k, j] = [j, k]$ for undirected graphs} \\
        = {} & \frac{1}{2}\sum_{j=1}^n\varphi([k, j])\phi([k, j])\left(g([j, k]) - g([k, j])\right).
    \end{align*}
    
    Note that
    \begin{align*}
        \langle \sI_{\cdot = k}, \divg g \rangle_\gV = {} & \sum_{i=1}^n(\sI_{i = k})(\divg g)(i)\chi(i) \notag \\
        = {} & (\divg g)(k)\chi(k).
    \end{align*}
    
    Note also that
    \begin{align*}
        {} & \langle \sI_{\cdot = k}, -\divg g \rangle_\gV = \langle \nabla\sI_{\cdot = k}, g \rangle_\gE \\
        \Longrightarrow {} & (\divg g)(k)\chi(k) = -\frac{1}{2}\sum_{j=1}^n\varphi([k, j])\phi([k, j])\left(g([j, k]) - g([k, j])\right) \\
        \Longrightarrow {} & (\divg g)(k) = \frac{1}{2\chi(k)}\sum_{j=1}^n\varphi([k, j])\phi([k, j])\left(g([k, j]) - g([j, k])\right).
    \end{align*}
\end{proof}
\subsubsection{Proof of \cref{thrm:lem2}}\label{app:thrm:lem2}
\textbf{\cref{thrm:lem2}.}\textit{
    Given an undirected graph $\gG = (\gV, \gE)$ and a function $f: \gV \mapsto \R$, we have for all $i \in \gV$,
    \begin{equation*}
        (\Delta_\Phi f)(i) = \frac{1}{\chi(i)}\sum_{j=1}^n\varphi([i, j])^2\phi([i, j])\left(f(i) - f(j)\right).
    \end{equation*}
}
\begin{proof}
    For all $i \in \gV$,
    \begin{align*}
        (\Delta_\Phi f)(i) = {} & (-\divg\nabla f)(i) \\
        = {} & -\frac{1}{2\chi(i)}\sum_{j=1}^n\varphi([i, j])\phi([i, j])\left(\nabla f([i, j]) - \nabla f([j, i])\right) \\
        = {} & \frac{1}{2\chi(i)}\sum_{j=1}^n\varphi([i, j])\phi([i, j])\left(\nabla f([j, i]) - \nabla f([i, j])\right) \\
        = {} & \frac{1}{2\chi(i)}\sum_{j=1}^n\varphi([i, j])\phi([i, j])\left(\varphi([j, i])\left(f(i) - f(j)\right) - \varphi([i, j])\left(f(j) - f(i)\right)\right) \\
        = {} & \frac{1}{2\chi(i)}\sum_{j=1}^n\varphi([i, j])\phi([i, j])\left(\varphi([i, j])\left(f(i) - f(j)\right) - \varphi([i, j])\left(f(j) - f(i)\right)\right) \tag{$[j, i] = [i, j]$} \\
        = {} & \frac{1}{\chi(i)}\sum_{j=1}^n\varphi([i, j])^2\phi([i, j])\left(f(i) - f(j)\right).
    \end{align*}
\end{proof}

\subsubsection{\cref{thrm:prop1} and proof}\label{app:thrm:prop1}
\begin{proposition}\label[proposition]{thrm:prop1}
    For any graph $\gG = (\gV, \gE)$ and any positive real-valued functions $\chi: \gV \mapsto \R_+, \phi: \gE \mapsto \R^+, \varphi: \gE \mapsto \R_+^*$, the parameterized graph Laplacian $\Delta_\Phi: \gH(\gV, \chi) \mapsto \gH(\gV, \chi)$ is self-adjoint and positive semi-definite.
\end{proposition}
\begin{proof}
    By the definition of parameterized graph Laplacian,
    \begin{align*}
        \langle f, \Delta_\Phi g \rangle_\gV = {} & \langle f, -\divg\nabla g \rangle_\gV \\
        = {} & \langle \nabla f, \nabla g \rangle_\gE \\
        = {} & \langle -\divg\nabla f, g \rangle_\gV \\
        = {} & \langle \Delta_\Phi f, g \rangle_\gV,
    \end{align*}
    which indicates the self-adjoint and
    \begin{align*}
        \langle f, \Delta_\Phi f \rangle_\gV = {} & \langle f, -\divg\nabla f \rangle_\gV \\
        = {} & \langle \nabla f, \nabla f \rangle_\gE \geq 0,
    \end{align*}
    which indicates the positive semi-definite.
\end{proof}
\subsubsection{\cref{thrm:lem3} and proof}\label{app:thrm:lem3}

\begin{lemma}\label[lemma]{thrm:lem3}
    Given a vector function $f: \gV \mapsto \R^c$, for any positive real-valued functions $\chi: \gV \mapsto \R_+, \phi: \gE \mapsto \R_+, \varphi: \gE \mapsto \R_+^*$, we have
    \begin{equation}
        \left.\frac{\od\gS(f)}{\od f}\right|_i = 2\chi(i)(\Delta_\Phi f)(i).
    \end{equation}
\end{lemma}
\begin{proof}
    \begin{align*}
        \left.\frac{\od\gS(f)}{\od f}\right|_i = {} & \left.\frac{\od}{\od f}\frac{1}{2}\sum_{i=1}^n\sum_{j=1}^n\varphi([i, j])^2\phi([i, j])\left\|f(j) - f(i)\right\|^2\right|_i \\
        = {} & \sum_{i=1}^n\varphi([i, j])^2\phi([i, j])(f(j) - f(i)) + \sum_{j=1}^n\varphi([i, j])^2\phi([i, j])(f(i) - f(j)) \\
        = {} & \sum_{j=1}^n\varphi([j, i])^2\phi([j, i])(f(i) - f(j)) + \sum_{j=1}^n\varphi([i, j])^2\phi([i, j])(f(i) - f(j)) \tag{exchange $i$ and $j$ for the first term} \\
        = {} & 2\sum_{j=1}^n\varphi([i, j])^2\phi([i, j])(f(i) - f(j)) \tag{$[j, i] = [i, j]$} \\
        = {} & 2\chi(i)(\Delta_\Phi f)(i).
    \end{align*}
\end{proof}

\subsection{Theorems and proofs in \cref{sec:analysis}}

\subsubsection{Proof of \cref{thrm:lem4}}\label{app:thrm:lem4}

\textbf{\cref{thrm:lem4}} (Convergence and Uniqueness Analysis).\textit{
    Let $\gamma_{\text{max}}$ be the largest singular value of $\mC$. The fixed-point equilibrium equation (\cref{eq:equi}) has a unique solution if $\gamma_{\text{max}} < 1$. We can iterate the equation:
    \begin{equation*}
        f^{(t+1)} = \mY' + \mC f^{(t)}, \quad t = 0, 1, \dots.
    \end{equation*}
    to obtain the optimal solution $f^\star = \lim_{t \rightarrow \infty}f^{(t)}$ which is independent of the initial state $f^{(0)}$. 
}
\begin{proof}
    For $t \geq 1$ and the sequence $f^{(0)}, f^{(1)}, \dots$ generated by $f^{(t+1)} = \mY' + \mC f^{(t)}$,
    \begin{align*}
        \|f^{(t+1)} - f^{(t)}\| \leq {} & \left\|\mY' + \mC f^{(t)} - \mY' - \mC f^{(t-1)}\right\| \\
        = {} & \left\|(\mC f^{(t)} - \mC f^{(t-1)}\right\| \\
        \leq {} & \|\mC\|\|f^{(t)} - f^{(t-1)}\|.
    \end{align*}
    For $n > m \geq 1$, the following inequality holds,
    \begin{align*}
        \|f^{(n)} - f^{(m)}\| = {} & \|f^{(n)} - f^{(n-1)} + f^{(n-1)} - \dots + f^{(m+1)} - f^{(m)}\| \\
        \leq {} & \sum_{t=m}^{n-1}\|f^{(t+1)} - f^{(t)}\| \\
        \leq {} & \sum_{t=m}^{n-1}\|\mC\|^{t-m}\|f^{(m+1)} - f^{(m)}\| \\
        \leq {} & \sum_{t=m}^{n-1}\|\mC\|^{t-m}\|\mC\|^m\|f^{(1)} - f^{(0)}\| \\
        = {} & \|\mC\|^m\sum_{t=0}^{n-m-1}\|\mC\|^t\|f^{(1)} - f^{(0)}\| \\
        \leq {} & \|\mC\|^m\sum_{t=0}^\infty\|\mC\|^t\|f^{(1)} - f^{(0)}\| \\
        = {} & \|\mC\|^m(1 - \|\mC\|)^{-1}\|f^{(1)} - f^{(0)}\|.
    \end{align*}
    
    If $\gamma_{\text{max}} < 1$, we have $\|\mC\| = \gamma_{\text{max}} < 1$, which implies that $\|\mC\|^m(1 - \|\mC\|)^{-1}\|f^{(1)} - f^{(0)}\| \geq 0$, then the inequality $\|f^{(n)} - f^{(m)}\| \leq (\|\mC\|)^m(1 - \|\mC\|)^{-1}\|f^{(1)} - f^{(0)}\|$ is well-posed. Note also that when $\|\mC\| < 1$, we have $\lim_{t \rightarrow \infty}(\|\mC\|)^t = 0$ and $\lim_{t \rightarrow \infty}(\|\mC\|)^t(1 - \|\mC\|)^{-1} = 0$. Then the sequence $f^{(0)}, f^{(1)}, f^{(2)}, \dots$ is a Cauchy sequence because
    \begin{equation*}
        0 \leq \lim_{m \rightarrow \infty}\|f^{(n)} - f^{(m)}\| \leq \lim_{m \rightarrow \infty}(\|\mC\|)^m(1 - \|\mC\|)^{-1}\|f^{(1)} - f^{(0)}\| = 0.
    \end{equation*}
    Thus the sequence converges to some solution of $f^\star = \mY' + \mC f^\star$. 
    
    To prove the uniqueness, suppose that both $f^{(a)}$ and $f^{(b)}$ satisfy $f = \mY' + \mC f$, i.e., $f^{(a)} = \mY' + \mC f^{(a)}$ and $f^{(b)} = \mY' + \mC f^{(b)}$. Then we have
    \begin{align*}
        0 \leq {} & \|f^{(a)} - f^{(b)}\| \\
        = {} & \left\|\mY' + \mC f^{(a)} - \mY'(i) - \mC f^{(b)}\right\| \\
        = {} & \left\|\mC f^{(a)} - \mC f^{(b)}\right\| \\
        \leq {} & \|\mC\|\|f^{(a)} - f^{(b)}\| \\
        = {} & \|\mC\|\left\|\mY' + \mC f^{(a)} - \mY' - \mC f^{(b)}\right\| \\
        \leq {} & \lim_{t \rightarrow \infty}\left(\|\mC\|\right)^t\|f^{(a)} - f^{(b)}\| = 0, \tag{$\lim_{t \rightarrow \infty}\left(\|\mC\|\right)^t = 0$ when $\gamma_{\text{max}} < 1$}
    \end{align*}
    which indicates that $f^{(a)} = f^{(b)}$ and there exists unique solution to $f = \mY' + \mC f$ if $\gamma_{\text{max}} < 1$.
\end{proof}

\subsubsection{Proof of \cref{thrm:corol1}}\label{app:thrm:corol1}
\textbf{\cref{thrm:corol1}} (Over-Smoothing during Training (OST) Condition).\textit{
    Assume that the initial state $f^{(0)}$ at node $i$ is initialized as a function of $\vx_i$. Assume further that $\chi, \phi, \varphi$ are not functions of $\vx_i$, e.g., $\Delta$ is the unnormalized Laplacian or random walk Laplacian. Let $\gamma_{\text{max}}$ be the largest singular value of $\mC$. Then, if $\gamma_{\text{max}} < 1$, the equilibrium of the fixed point equation \cref{eq:equi} is independent of the node features $\vx_i$, i.e. the same equilibrium solution is obtained even if the node features $\vx_i$ are changed.
}
\begin{proof}
    When $\gamma_{\text{max}} < 1$, by \cref{thrm:lem4} we know that the fixed-point equation~\cref{eq:equi} has a unique equilibrium. Since $\chi, \varphi$, and $\varphi$ are not functions of $\vx_i$, $\Delta$ is independent of the node features $\vx_i, \forall i \in \gV$. Therefore, $\gamma_{\text{max}}$ is independent of the node features, which indicates that the uniqueness condition of the equilibrium is also independent of the node features.
\end{proof}

\subsubsection{Proof of \cref{thrm:lem5}}\label{app:thrm:lem5}
\textbf{\cref{thrm:lem5}} (Over-Smoothing during Inference (OSI) Condition).\textit{
    When $\gI = \emptyset$, i.e., there is no constrained nodes in \cref{eq:Dirichlet}, we have $\mC = \mP$ and $\rvy'(i) = \mathbf{0}, \forall i \in \gV$. $\mP$ is a Markov matrix and the solution of \cref{eq:Dirichlet} satisfies $f = \mP f$. If the graph $\gG$ is connected and non-bipartite, i,e., the Markov chain corresponding to $\mP$ is irreducible and aperiodic, the equilibrium $f^\star(i)$ for all nodes are identical and not unique, i.e., $f^\star(i) = \vv, \forall i \in \gV$, where $\vv$ is any vector in $\R^c$ and is independent of the labels. Iterating $f^{(t+1)} = \mP f^{(t)}, t = 0, 1, \dots,$ with initial state $f^{(0)} \in \R^{n \times c}$ will converge to
    \begin{equation*}
        f^\star(i) = (\lim_{t \rightarrow \infty}\mP^tf^{(0)})(i) = (\vpi f^{(0)})^\top, \quad \forall i \in \gV,
    \end{equation*}
    where $\boldsymbol{\pi} = (\widehat{\emD}_1 / \sum_{j}\widehat{\emD}_j, \dots, \widehat{\emD}_n / \sum_{j}\widehat{\emD}_j) \in \R^{1 \times n}$ is the stationary distribution of $\mP$.
}
\begin{proof}
    If the graph is connected and non-bipartite, the Markov chain corresponding to $\mP$ is irreducible and aperiodic. Then by the Fundamental Theorem of Markov chains~\citep{Markov-Book2012}, $\mP$ has a unique stationary distribution. Let $\vpi = (\pi_1, \dots, \pi_n) \in \R^{1 \times n}$ be the stationary distribution of $\mP$, then
    \begin{equation*}
        \vpi = \vpi\mP.
    \end{equation*}
    $\vpi$ is given by $(\widehat{\emD}_1 / \sum_{j}\widehat{\emD}_j, \dots, \widehat{\emD}_n / \sum_{j}\widehat{\emD}_j)$, which can be easily verified by
    \begin{align*}
        \sum_{i=1}^n\pi_i\emP_{i,j} = {} & \sum_{i=1}^n\frac{\widehat{\emD}_i}{\sum_{k=1}^n\widehat{\emD}_k}\frac{\varphi([i, j])^2\phi([i, j])}{\widehat{\emD}_i} \\
        = {} & \frac{\sum_{i=1}^n\varphi([i, j])^2\phi([i, j])}{\sum_{k=1}^n\widehat{\emD}_k} \\
        = {} & \frac{\widehat{\emD}_j}{\sum_{k=1}^n\widehat{\emD}_k} \\
        = {} & \pi_j.
    \end{align*}
    By the Fundamental Theorem of Markov chains, we have
    \begin{equation*}
        \lim_{t \rightarrow \infty}(\mP^t)_{i,j} = \pi_j, \quad \text{for all } i, j \in \gV,
    \end{equation*}
    which indicates that as $t$ goes to infinity, all row vectors in $\mP^t$ are identical and is the stationary vector $\vpi$ of $\mP$, i.e., $\lim_{t \rightarrow \infty}\mP^t = (\vpi^\top, \dots, \vpi^\top)^\top$. Denote $f^\star = (f^\star(1), \dots, f^\star(n))^\top \in \R^{n \times c}$, then
    \begin{equation*}
        f^\star(i) = (\mP f^\star)(i) = (\mP\mP f^\star)(i) = (\lim_{t \rightarrow \infty}\mP^tf^\star)(i) = ((\vpi^\top, \dots, \vpi^\top)^\top f)(i) = (\vpi f^\star)^\top, \quad \forall i \in \gV.
    \end{equation*}
    which indicates that the equilibrium of all nodes are identical. $(\vpi f)^\top$ can be any vector in $\R^c$, therefore any vector $\vv \in \R^c$, $f^\star = (\vv, \dots, \vv)^\top \in \R^{n \times d}$ is an equilibrium of the fixed-point equation $f = \mP f$. Hence, the equilibrium is a constant function over all nodes.
    
    Moreover, iterating $f^{(t+1)} = \mP f^{(t)}$ with an initial state $f^{(0)}$ with converge to
    \begin{align*}
        \lim_{t \rightarrow \infty}f^{(t+1)}(i) = (\lim_{t \rightarrow \infty}\mP^tf^{(0)})(i) = ((\vpi^\top, \dots, \vpi^\top)^\top f^{(0)})(i) = (\vpi f^{(0)})^\top, \quad \text{for all } i \in \gV,
    \end{align*}
    which is an equilibrium of the fixed-point equation $f = \mP f$.
\end{proof}

\subsection{Theorems and proofs in \cref{sec:model}}

\subsubsection{Proof of \cref{thrm:thrm1}}\label{app:thrm:thrm1}
\textbf{\cref{thrm:thrm1}} (Spectral Range of $\Delta_\Phi$).\textit{
    Let $\Delta_\Phi$ be the graph neural Laplacian and $\chi, \phi, \varphi$ are given by \cref{eq:eq12,eq:eq13,eq:eq14} respectively, $\lambda_\Phi$ be an eigenvalue associated with the eigenvector $\vu$ of $\Delta_\Phi$. Suppose \cref{assump:graph,assump:embedding,assump:param} hold, then
    \begin{equation*}
        0 \leq \lambda_\Phi \leq 2c_\phi^2c_\chi c_X\cosh(c_\chi c_X).
    \end{equation*}
}
\begin{proof}
    Note that by \cref{thrm:prop1} the graph neural Laplacian is positive semi-definite and therefore we have $\lambda_\Phi \geq 0$. On the other hand, by the definition of parameterized graph Laplacian, we have for all $i = 1, 2, \dots, n$,
    \begin{equation*}
        (\Delta_\Phi\vu)_i = \frac{1}{\chi(i)}\sum_{j=1}^n\varphi([i, j])^2\phi([i, j])(\evu_i - \evu_j) = \lambda_\Phi\evu_i.
    \end{equation*}

    Then we have for all $i = 1, \dots, n$,
    \begin{align*}
        \lambda_\Phi = {} & \frac{1}{\evu_i}\sum_{j=1}^n\frac{\varphi([i, j])^2\phi([i, j])}{\chi(i)}(\evu_i - \evu_j) \\
        = {} & \sum_{j=1}^n\frac{\varphi([i, j])^2\phi([i, j])}{\chi(i)}(1 - \frac{\evu_j}{\evu_i}) \\
        \leq {} & \sum_{j=1}^n\frac{\varphi([i, j])^2\phi([i, j])}{\chi(i)}\left(1 + \left|\frac{\evu_j}{\evu_i}\right|\right) \\
        = {} & \sum_{j=1}^n\frac{\emA_{i,j}}{\emD_i}\frac{\tanh((\|\Theta_\varphi(\vx_i - \vx_j)\| + \epsilon)^{-1})\tanh(|(\Theta_\phi\Theta_\chi\vx_i)^\top(\Theta_\phi\Theta_\chi\vx_j)|)}{\tanh(\|\Theta_\chi\vx_i\|)}\left(1 + \left|\frac{\evu_j}{\evu_i}\right|\right) \\
        \leq {} & \sum_{j=1}^n\frac{\emA_{i,j}}{\emD_i}\frac{\tanh(|(\Theta_\phi\Theta_\chi\vx_i)^\top(\Theta_\phi\Theta_\chi\vx_j)|)}{\tanh(\|\Theta_\chi\vx_i\|)}\left(1 + \left|\frac{\evu_j}{\evu_i}\right|\right) \tag{$\tanh(a) \leq 1, \forall a \in \R$} \\
        \leq {} & \sum_{j=1}^n\frac{\emA_{i,j}}{\emD_i}\frac{|(\Theta_\phi\Theta_\chi\vx_i)^\top(\Theta_\phi\Theta_\chi\vx_j)|}{\tanh(\|\Theta_\chi\vx_i\|)}\left(1 + \left|\frac{\evu_j}{\evu_i}\right|\right) \tag{$\tanh(a) \leq a, \forall a \in \R$} \\
        = {} & \sum_{j=1}^n\frac{\emA_{i,j}}{\emD_i}\frac{|(\Theta_\phi\Theta_\chi\vx_i)^\top(\Theta_\phi\Theta_\chi\vx_j)|\cosh(\|\Theta_\chi\vx_i\|)}{\sinh(\|\Theta_\chi\vx_i\|)}\left(1 + \left|\frac{\evu_j}{\evu_i}\right|\right) \\
        \leq {} & \sum_{j=1}^n\frac{\emA_{i,j}}{\emD_i}\frac{\|\Theta_\phi\|\|\Theta_\chi\vx_i\|\|\Theta_\phi\Theta_\chi\vx_j\|\cosh(\|\Theta_\chi\vx_i\|)}{\sinh(\|\Theta_\chi\vx_i\|)}\left(1 + \left|\frac{\evu_j}{\evu_i}\right|\right) \\
        \leq {} & \sum_{j=1}^n\frac{\emA_{i,j}}{\emD_i}\|\Theta_\phi\|\|\Theta_\phi\Theta_\chi\vx_j\|\cosh(\|\Theta_\chi\vx_i\|)\left(1 + \left|\frac{\evu_j}{\evu_i}\right|\right) \tag{$\frac{a}{\sinh(a)} \leq 1, \forall a \in \R$} \\
        \leq {} & \sum_{j=1}^n\frac{\emA_{i,j}}{\emD_i}\|\Theta_\phi\|^2\|\Theta_\chi\|\|\vx_j\|\cosh(\|\Theta_\chi\|\|\vx_i\|)\left(1 + \left|\frac{\evu_j}{\evu_i}\right|\right) \\
        \leq {} &  c_\phi^2c_\chi c_X\cosh(c_\chi c_X)\left(\sum_{j=1}^n\frac{\emA_{i,j}}{\emD_i}\left(1 + \left|\frac{\evu_j}{\evu_i}\right|\right)\right).
    \end{align*}
    Since the above inequality holds for all $i = 1, \dots, n$, let $k = \argmax_i(\{|\evu_i|\}_{i=1, 2, \dots, n})$, we have
    \begin{align*}
        \lambda_\Phi \leq {} & c_\phi^2c_\chi c_X\cosh(c_\chi c_X)\left(\sum_{j=1}^n\frac{\emA_{k,j}}{\emD_k}\left(1 + \left|\frac{\evu_j}{\evu_k}\right|\right)\right) \\
        \leq {} & c_\phi^2c_\chi c_X\cosh(c_\chi c_X)\left(\sum_{j=1}^n\frac{\emA_{k,j}}{\emD_k}\left(1 + 1\right)\right) \\
        = {} & 2c_\phi^2c_\chi c_X\cosh(c_\chi c_X).
    \end{align*}
\end{proof}

\subsubsection{Proof of \cref{thrm:corol3}}\label{app:thrm:corol3}
\textbf{\cref{thrm:corol3}} (Tractable Well-Posedness Condition and Convergence Rate).\textit{
    Given an undirected graph $\gG = (\gV, \gE, \mA)$ with node embeddings $\widetilde{\mX} := (\widetilde{\vx}_1, \dots, \widetilde{\vx}_n)^\top \in \R^{n \times d}$, Let $\lambda_{\text{max}}$ be the largest eigenvalues of the matrix $\Delta$. Suppose that \cref{assump:graph} holds, then, the fixed-point equilibrium equation $\mZ = \widetilde{\mX} - \frac{1}{\mu}\Delta\mZ$ has a unique solution if $\mu > \lambda_{\text{max}}$. The solution can be obtained by iterating:
    \begin{equation*}
        \mZ^{(t+1)} = \widetilde{\mX} - \frac{1}{\mu}\Delta\mZ^{(t)}, \text{ with } \mZ^{(0)} = \mathbf{0}, t = 0, 1, \dots.
    \end{equation*}
    Therefore, $\mZ^\star = \lim_{t \rightarrow \infty}\mZ^{(t)}$. Suppose that $\|\mZ^\star\| \leq c_Z \in \R_+^*$, then $\forall t \geq 1$, $\|\mZ^{(t)} - \mZ^\star\| \leq c_Z\left(\lambda_{\text{max}} / \mu\right)^t$. Specifically,}
    \begin{enumerate}
        \item \textit{If $\Delta$ is the random walk Laplacian, i.e., $\Delta \equiv \Delta^{\text{(rw)}}$, $\mu > 2$ and $\|\mZ^{(t)} - \mZ^\star\| \leq c_Z\left(2 / \mu\right)^t$;}

        \item \textit{If $\Delta$ is learnable and $\Delta \equiv \Delta_\Phi$, suppose further that \cref{assump:embedding,assump:param} hold, then we have $\mu > 2c_\phi^2c_\chi c_X\cosh(c_\chi c_X)$ and $\|\mZ^{(t)} - \mZ^\star\| \leq c_Z(2c_\phi^2c_\chi c_X\cosh(c_\chi c_X) / \mu)^t$.}
    \end{enumerate}
\begin{proof}
    The fixed-point equilibrium equation $\mZ = \widetilde{\mX} - \frac{1}{\mu}\Delta\mZ$ is a case of $f = \mY' + \mC f$ when $\gI \equiv \gV$. Then we have $\mC = -\frac{1}{\mu}\Delta$. By \cref{thrm:prop1}, $\Delta$ is positive semi-definite. Therefore, the largest singular value of $\Delta$ is its largest eigenvalue. Then we have
    \begin{equation*}
        \|\mC\| = \|\frac{1}{\mu}\Delta\| = \frac{1}{\mu}\|\Delta\| = \frac{1}{\mu}\max_i|\lambda_i| = \frac{1}{\mu}\lambda_{\text{max}}, 
    \end{equation*}
    which indicates that $\gamma = \frac{1}{\mu}\lambda_{\text{max}}$.
    By \cref{thrm:lem4}, we have if $\frac{1}{\mu}\lambda_{\text{max}} < 1$, i.e., $\mu > \lambda_{\text{max}}$, the fixed-point equilibrium equation has a unique solution and iterating \cref{eq:iter} will convergence to the solution, i.e., $\mZ^\star = \lim_{t \rightarrow \infty}\mZ^{(t)}$.
    Moreover, for $t \geq 1$,
    \begin{align*}
        \|\mZ^{(t)} - \mZ^\star\| = {} & \|\widetilde{\mX} - \frac{1}{\mu}\Delta\mZ^{(t-1)} - \widetilde{\mX} + \frac{1}{\mu}\Delta\mZ^\star\| \\
        = {} & \frac{1}{\mu}\|\Delta(\mZ^{(t-1)} - \mZ^\star)\| \\
        \leq {} & \frac{1}{\mu}\|\Delta\|\|\mZ^{(t-1)} - \mZ^\star\| \\
        \leq {} & \left(\frac{1}{\mu}\|\Delta\|\right)^{t}\|\mZ^{(0)} - \mZ^\star\| \\
        = {} & \left(\frac{\lambda_{\text{max}}}{\mu}\right)^{t}\|\mZ^\star\| \\
        \leq {} & c_Z\left(\frac{\lambda_{\text{max}}}{\mu}\right)^{t}.
    \end{align*}
    
    \textbf{1. When $\Delta$ is fixed and $\Delta \equiv \Delta^{\text{(rw)}}$.} The largest eigenvalue of the random walk matrix $\lambda_{\text{max}} = 2$. Therefore, we have $\mu > 2$ to ensure the uniqueness of the fixed-point equation and the convergence of \cref{eq:iter}. 

    \textbf{2. When $\Delta$ is parameterized and $\Delta \equiv \Delta_\Phi$.} With \cref{assump:embedding,assump:param}, $\|\widetilde{\vx}_i\| \leq c_X, \forall i \in \gV, \|\Theta_\chi\| \leq c_\chi, \|\Theta_\phi\| \leq c_\phi$, the largest eigenvalue of $\Delta_\Phi$ is upper-bounded by $\lambda_{\text{max}} \leq 2c_\phi^2c_\chi c_X\cosh(c_\chi c_X)$ by \cref{thrm:thrm1}. Therefore, to ensure the uniqueness of the fixed-point equation and the convergence of \cref{eq:iter} we need $2c_\phi^2c_\chi c_X\cosh(c_\chi c_X) < \mu$. 
\end{proof}

\subsubsection{Proof of \cref{thrm:dignn-generalization}}

\textbf{Transductive learning on graphs.} 
Given an undirected graph $\gG = (\gV, \gE, \mA)$ with node features and labels $\{(\vx_i, y_i)\}_{i=1}^{m+u}$ where $m+u = n$. 
Without loss of generality, let $\{y_i\}_{i=1}^m$ be the selected node labels, we aim to predict the labels of all nodes by a learner (model) trained on the graph $\gG$ with $\{\vx_i\}_{i=1}^{m+u} \cup \{y_i\}_{i=1}^m$. 
For any $f \in \gH$, the training and test error is defined as: 
\begin{align*}
    \widehat{L}_m(f) := {} & \frac{1}{m}\sum_{i=1}^m\ell(f(\vx_i), y_i), \\
    L_u(f) := {} & \frac{1}{u}\sum_{i=m+1}^{m+u}\ell(f(\vx_i), y_i)
\end{align*} 
respectively, where $\ell: \gH \times \gX \times \gY \mapsto \R_+$ is the loss function. We are interested in the transductive generalization gap of $f$ which is defined as $|\widehat{L}_m(f) - L_u(f)|$.

\begin{definition}[Rademacher Complexity~\citep{book/shalev2014understanding,book/mohri2018foundations}]
    For a class $\gH \subseteq \R^{\gX}$, a sample $S = \{\vx_1, \dots, \vx_n\}$ of size $n$ drawn i.i.d. from a distribution $\gD$, the (empirical) Rademacher complexity of $\gH$ w.r.t the sample $S$:
    \begin{equation*}
        \Re_n(\gH) := \E_{\vvarepsilon}\left[\sup_{h \in \gH}\frac{1}{n}\sum_{i=1}^n\varepsilon_ih(x_i)\right],
    \end{equation*}
    where $\vvarepsilon = (\varepsilon_1, \dots, \varepsilon_n)$ are the Rademacher random variables with $\Pr(\varepsilon_i = -1) = \Pr(\varepsilon_i = +1)$. 
\end{definition}

\begin{definition}[Transductive Rademacher Complexity~\citep{jair/El-YanivP09}]
    Let $\gA \subseteq \R^{m+u}$ and $p \in [0, 1/2]$. Let $\vsigma = (\sigma_1, \dots, \sigma_{m+u})^\top$ be a vector of i.i.d. random variables such that
    \begin{equation}
        \sigma_i :=
        \begin{cases}
            1 & \text{ with probability } p; \\
            -1 & \text{ with probability } p; \\
            0 & \text{ with probability } 1-2p.
        \end{cases}
    \end{equation}
    The transductive Rademacher complexity with parameter $p$ is defined as
    \begin{equation*}
        \widetilde{\Re}_{m+u}(\gA, p) := \left(\frac{1}{m} + \frac{1}{\mu}\right) \cdot \E_{\vsigma}\left[\sup_{\va \in \gA}\sum_{i=1}^{m+u}\sigma_i\eva_i\right].
    \end{equation*}

    Specially, we abbreviate $\widetilde{\Re}_{m+u}(\gA) := \widetilde{\Re}_{m+u}(\gA, p_0)$ with $p_0 := \frac{mu}{(m+u)^2}$.
\end{definition}

\begin{lemma}[\citet{jair/El-YanivP09}]\label{app:thrm:compare_rad}
    For any $\gA \subseteq \R^{m+u}$ and $0 \leq p_1 < p_2 \leq 1/2$,
    \begin{equation*}
        \widetilde{\Re}_{m+u}(\gA, p_1) < \widetilde{\Re}_{m+u}(\gA, p_2).
    \end{equation*}
\end{lemma}

\begin{theorem}[Transductive Generalization Bound~\citep{jair/El-YanivP09}]\label{app:thrm:transductive}
    Let $c_0 := \sqrt{\frac{32\ln(4e)}{3}} < 5.05$ and $c_1 := (\frac{1}{m} + \frac{1}{u}), c_2 := \frac{m+u}{(m+u-1/2)(1-1/2(\max(m, u)))}$. Then, for any $\delta > 0$, with probability of at least $1-\delta$ over the choice of the training set from $S$, for all $h \in \gH$,
    \begin{equation*}
        L_u(h) \leq \widehat{L}_m(h) + \widetilde{\Re}_{m+u}(\ell \circ \gH) + c_0c_1\sqrt{\min(m, u)} + \sqrt{\frac{c_1c_2}{2}\ln\frac{1}{\delta}}.
    \end{equation*}
\end{theorem}

\begin{lemma}[Talagrand’s Contraction Lemma~\citep{book/ledoux1991}]\label{app:thrm:contract}
    For each $i \in [n]$, let $\phi_i: \R \rightarrow \R$ be a $\rho$-Lipschitz function, that is,
    \begin{equation*}
        |\phi_i(x) - \phi_i(y)| \leq \rho|x - y|, \quad \forall x, y \in \R, \forall i \in [n].
    \end{equation*}
    For a vector $\vv \in \R^n$, let $\Phi(\vv) = (\phi_1(\evv_1), \dots,  \phi_n(\evv_n))$ and for a set $\gA$, let $\Phi \circ \gA = \{\Phi(\va) \mid \va \in \gA\}$. Then
    \begin{equation*}
        \Re_n(\Phi \circ \gA) = \rho \cdot \Re_n(\gA).
    \end{equation*}
\end{lemma}

\begin{lemma}\label{app:thrm:transductive2rade}
    For any $\gA \subseteq \R^{m+u}$ with $m+u=n$, we have
    \begin{equation*}
        \widetilde{\Re}_{m+u}(\gA) < \Re_n(\gA). 
    \end{equation*}
\end{lemma}
\begin{proof}
    By \cref{app:thrm:compare_rad}, for all $p_0 := \frac{mu}{(m+u)^2} < 1/2$, we have
    \begin{align*}
        \widetilde{\Re}_{m+u}(\gA) = {} & \widetilde{\Re}_{m+u}(\gA, p_0) \\
        < {} & \widetilde{\Re}_{m+u}(\gA, 1/2) \\
        = {} & \Re_n(\gA).
    \end{align*}
\end{proof}

\begin{lemma}[Rademacher Complexity of Implicit Graph Neural Diffusion Layers]\label{app:thrm:imp-rade}
    Given a graph $\gG$ with $n$ vertices and the adjacency matrix $\mA \in \R_+^{n \times n}$ and node embeddings $\widetilde{\mX} = (\widetilde{\vx}_1, \dots, \widetilde{\vx}_n)^\top \in \R^{n \times d}$ with \cref{assump:embedding} holds, i.e., $\|\widetilde{\vx}_i\|_2 \leq c_X$. Let $\Delta_\Phi$ be defined by functions $\chi, \phi, \varphi$ as given in \cref{eq:eq12,eq:eq13,eq:eq14} with learnable parameters $\Theta_\chi \cup \Theta_\phi \cup \Theta_\varphi =: \Phi \in \Omega$, $\lambda_\Phi$ be the largest eigenvalue of $\Delta_\Phi$. For any integer $T \geq 1$ and $\mZ^{(0)} = \mathbf{0}$, define the class of functions $\gF_T$ by:
    \begin{equation*}
        \gF_T := \{\widetilde{\mX} \mapsto \mZ^{(t+1)} = \widetilde{\mX} - \frac{1}{\mu}\Delta_\Phi\mZ^{(t)}: \Phi \in \Omega, t = 0, 1, \dots, T-1\}.
    \end{equation*}
    Then, we have
    \begin{equation*}
        \Re_n\left(\gF_T\right) := \frac{1}{n}\E_{\vvarepsilon}\left[\sup_{f \in \gF_T}\left\|\vvarepsilon^\top f(\widetilde{\mX})\right\|\right] \leq \frac{c_X}{\sqrt{n}}\sum_{t=0}^{T-1}\left(\frac{\lambda_\Phi}{\mu}\right)^{t}.
    \end{equation*}

    Suppose further that \cref{assump:graph,assump:param} hold, we have
    \begin{equation*}
        \Re_n\left(\gF_T\right) \leq \frac{c_X}{\sqrt{n}}\sum_{t=0}^{T-1}\left(\frac{2c_\phi^2c_\chi c_X\cosh(c_\chi c_X)}{\mu}\right)^{t}.
    \end{equation*}
\end{lemma}
\begin{proof}
    Note that for any matrix $\mU := (\vu_1, \dots, \vu_n)^\top \in \R^{n \times d}$, we have
    \begin{align}
        \E_{\vvarepsilon}\left[\|\vvarepsilon^\top\mU\|\right] = {} & \E_{\vvarepsilon}\left[\sqrt{\|\vvarepsilon^\top\mU\|^2}\right] \leq \sqrt{\E_{\vvarepsilon}\left[\left\|\sum_{i=1}^n\varepsilon_i\vu_i\right\|^2\right]} \tag{Jensen's inequality} \\
        = {} & \sqrt{\E_{\vvarepsilon}\left[\sum_{i=1}^n\sum_{j=1}^n\varepsilon_i\varepsilon_j\vu_i^\top\vu_j\right]} = \sqrt{\sum_{i=1}^n\|\vu_i\|^2} = \|\mU\|_F. \label{eq:eq3}
    \end{align}
    
    Then, we have
    \begin{align*}
        n\Re_n\left(\gF_T\right) = {} & \E_{\vvarepsilon}\left[\sup_{f \in \gF_T}\left\|\vvarepsilon^\top f(\widetilde{\mX})\right\|\right] = \E_{\vvarepsilon}\left[\sup_{f \in \gF_{T-1}}\left\|\vvarepsilon^\top\left(\widetilde{\mX} - \frac{1}{\mu}\Delta_\Phi f(\widetilde{\mX})\right)\right\|\right] \\
        \leq {} & \E_{\vvarepsilon}\left[\left\|\vvarepsilon^\top\widetilde{\mX}\right\| + \sup_{\Phi \in \Omega, f \in \gF_{T-1}}\left\|\vvarepsilon^\top\frac{1}{\mu}\Delta_\Phi f(\widetilde{\mX})\right\|\right] \\
        \leq {} & \|\widetilde{\mX}\|_F + \E_{\vvarepsilon}\left[\sup_{\Phi \in \Omega, f \in \gF_{T-1}}\left\|\vvarepsilon^\top\frac{1}{\mu}\Delta_\Phi f(\widetilde{\mX})\right\|\right] \tag{by (\ref{eq:eq3})} \\
        = {} & \|\widetilde{\mX}\|_F + \E_{\vvarepsilon}\left[\sup_{\Phi \in \Omega, f \in \gF_{T-2}}\left\|\vvarepsilon^\top\frac{1}{\mu}\Delta_\Phi\left(\widetilde{\mX} - \frac{1}{\mu}\Delta_\Phi f(\widetilde{\mX})\right)\right\|\right] \\
        \leq {} & \|\widetilde{\mX}\|_F + \E_{\vvarepsilon}\left[\sup_{\Phi \in \Omega}\left\|\vvarepsilon^\top\frac{1}{\mu}\Delta_\Phi\widetilde{\mX}\right\| + \sup_{\Phi \in \Omega, f \in \gF_{T-2}}\left\|\vvarepsilon^\top\left(\frac{1}{\mu}\Delta_\Phi\right)^2f(\widetilde{\mX})\right\|\right] \\
        \leq {} & \|\widetilde{\mX}\|_F + \sup_{\Phi \in \Omega}\frac{1}{\mu}\left\|\Delta_\Phi\widetilde{\mX}\right\|_F + \E_{\vvarepsilon}\left[\sup_{\Phi \in \Omega, f \in \gF_{T-2}}\left\|\vvarepsilon^\top\left(\frac{1}{\mu}\Delta_\Phi\right)^2f(\widetilde{\mX})\right\|\right] \tag{by (\ref{eq:eq3})} \\
        \leq {} & \|\widetilde{\mX}\|_F + \sup_{\Phi \in \Omega}\frac{1}{\mu}\left\|\Delta_\Phi\right\|\|\widetilde{\mX}\|_F + \E_{\vvarepsilon}\left[\sup_{\Phi \in \Omega, f \in \gF_{T-2}}\left\|\vvarepsilon^\top\left(\frac{1}{\mu}\Delta_\Phi\right)^2f(\widetilde{\mX})\right\|\right] \tag{$\|AB\|_F \leq \|A\|\cdot\|B\|_F$} \\
        = {} & \sup_{\Phi \in \Omega}\|\widetilde{\mX}\|_F\sum_{t=0}^1\left(\frac{1}{\mu}\|\Delta_\Phi\|\right)^{t} + \E_{\vvarepsilon}\left[\sup_{\Phi \in \Omega, f \in \gF_{T-2}}\left\|\vvarepsilon^\top\left(\frac{1}{\mu}\Delta_\Phi\right)^2f(\widetilde{\mX})\right\|\right] \\
        \leq {} & \sup_{\Phi \in \Omega}\|\widetilde{\mX}\|_F\sum_{t=0}^{T-2}\left(\frac{1}{\mu}\|\Delta_\Phi\|\right)^{t} + \E_{\vvarepsilon}\left[\sup_{\Phi \in \Omega, f \in \gF_{1}}\left\|\vvarepsilon^\top\left(\frac{1}{\mu}\Delta_\Phi\right)^{T-1}f(\widetilde{\mX})\right\|\right] \\
        \leq {} & \sup_{\Phi \in \Omega}\|\widetilde{\mX}\|_F\sum_{t=0}^{T-2}\left(\frac{1}{\mu}\|\Delta_\Phi\|\right)^{t} + \E_{\vvarepsilon}\left[\sup_{\Phi \in \Omega}\left\|\vvarepsilon^\top\left(\frac{1}{\mu}\Delta_\Phi\right)^{T-1}\left(\widetilde{\mX} - \frac{1}{\mu}\Delta_\Phi\mZ^{(0)}\right)\right\|\right] \\
        = {} & \sup_{\Phi \in \Omega}\|\widetilde{\mX}\|_F\sum_{t=0}^{T-2}\left(\frac{1}{\mu}\|\Delta_\Phi\|\right)^{t} + \E_{\vvarepsilon}\left[\sup_{\Phi \in \Omega}\left\|\vvarepsilon^\top\left(\frac{1}{\mu}\Delta_\Phi\right)^{T-1}\widetilde{\mX}\right\|\right] \tag{$\mZ^{(0)} = \mathbf{0}$} \\
        \leq {} & \sup_{\Phi \in \Omega}\|\widetilde{\mX}\|_F\sum_{t=0}^{T-1}\left(\frac{1}{\mu}\|\Delta_\Phi\|\right)^{t}. \tag{by \ref{eq:eq3}}
    \end{align*}

    By \cref{assump:embedding}, $\|\widetilde{\vx}\|_2 \leq c_X$, we have $\|\widetilde{\mX}\| \leq \sqrt{n}c_X$. Then, by $\|\Delta_\Phi\| \leq \lambda_\Phi$, we have
    \begin{align*}
        \Re_n\left(\gF_T\right) \leq {} & \frac{1}{n}\sup_{\Phi \in \Omega}\|\widetilde{\mX}\|_F\sum_{t=0}^{T-1}\left(\frac{1}{\mu}\|\Delta_\Phi\|\right)^{t} \\
        \leq {} & \frac{c_X}{\sqrt{n}}\sum_{t=0}^{T-1}\left(\frac{\lambda_\Phi}{\mu}\right)^{t}.
    \end{align*}
    
    Furthermore, by \cref{assump:param}, $\|\Theta_\chi\| \leq c_\chi, \Theta_\phi \leq c_\phi$, and \cref{assump:graph}, we have $\lambda_{\max} \leq 2c_\phi^2c_\chi c_X\cosh(c_\chi c_X)$ in terms of \cref{thrm:thrm1}. Therefore, we obtain
    \begin{equation*}
        \Re_n\left(\gF_T\right) \leq \frac{c_X}{\sqrt{n}}\sum_{t=0}^{T-1}\left(\frac{2c_\phi^2c_\chi c_X\cosh(c_\chi c_X)}{\mu}\right)^{t}.
    \end{equation*}
\end{proof}

\begin{theorem}[Rademacher Complexity of DIGNN-$\Delta_\Phi$]\label{app:thrm:dignn-rade}
    Let $\Delta_\Phi$ be defined by functions $\chi, \phi, \varphi$ as given in \cref{eq:eq12,eq:eq13,eq:eq14} with learnable parameters $\Theta_\chi \cup \Theta_\phi \cup \Theta_\varphi =: \Phi \in \Omega$, $\lambda_\Phi$ be the largest eigenvalue of $\Delta_\Phi$.
    Let \cref{eq:model-in,eq:model-out,eq:iter} be the input, the output, and the implicit graph diffusion layers of DIGNN-$\Delta_\Phi$ respectively.
    Suppose that \cref{assump:loss,assump:output} hold and the node embeddings $\widetilde{\mX}$ of the first layer (\cref{eq:model-in}) of DIGNN-$\Delta_\Phi$ satisfy \cref{assump:embedding}. 
    Then, for any integer $T \geq 1$, we have
    \begin{equation*}
        \Re_n(\ell \circ h_{\Theta^{(2)}} \circ \gF_T) \leq \rho_\ell\rho_h\frac{c_X}{\sqrt{n}}\sum_{t=0}^{T-1}\left(\frac{\lambda_\Phi}{\mu}\right)^{t}.
    \end{equation*}

    Suppose further that \cref{assump:graph,assump:param} hold, we have
    \begin{equation*}
        \Re_n(\ell \circ h_{\Theta^{(2)}} \circ \gF_T) \leq \rho_\ell\rho_h\frac{c_X}{\sqrt{n}}\sum_{t=0}^{T-1}\left(\frac{2c_\phi^2c_\chi c_X\cosh(c_\chi c_X)}{\mu}\right)^{t}.
    \end{equation*}
\end{theorem}
\begin{proof}
    By \cref{assump:loss,assump:output}, we know that $\ell$ is $\rho_\ell$-Lipschitz continuous and $h_{\Theta^{(2)}}$ is $\rho_h$-Lipschitz continuous. Therefore, $\ell \circ h_{\Theta^{(2)}}$ is $\rho_\ell\rho_h$-Lipschitz continuous. Then, by Talagrand’s Contraction Lemma (\cref{app:thrm:contract}), we have
    \begin{equation*}
        \Re_n(\ell \circ h_{\Theta^{(2)}} \circ \gF_T) \leq \rho_\ell\rho_h \cdot \Re_n(\gF_T).
    \end{equation*}

    By \cref{assump:embedding} and \cref{app:thrm:imp-rade}, we obtain
    \begin{equation*}
        \Re_n(\ell \circ h_{\Theta^{(2)}} \circ \gF_T) \leq \rho_\ell\rho_h \cdot \Re_n(\gF_T) \leq \rho_\ell\rho_h\frac{c_X}{\sqrt{n}}\sum_{t=0}^{T-1}\left(\frac{\lambda_\Phi}{\mu}\right)^{t}.
    \end{equation*}

    Furthermore, if \cref{assump:graph,assump:param} hold, we have
    \begin{equation*}
        \Re_n(\ell \circ h_{\Theta^{(2)}} \circ \gF_T) \leq \rho_\ell\rho_h\frac{c_X}{\sqrt{n}}\sum_{t=0}^{T-1}\left(\frac{2c_\phi^2c_\chi c_X\cosh(c_\chi c_X)}{\mu}\right)^{t}.
    \end{equation*}
\end{proof}

Now we are ready to prove \cref{thrm:dignn-generalization}:

\textbf{\cref{thrm:dignn-generalization}} (Transductive Generalization Bounds for DIGNN-$\Delta_\Phi$).\textit{
    Let $\Delta_\Phi$ be defined by functions $\chi, \phi, \varphi$ as given in \cref{eq:eq12,eq:eq13,eq:eq14} with learnable parameters $\Theta_\chi \cup \Theta_\phi \cup \Theta_\varphi =: \Phi \in \Omega$, $\lambda_\Phi$ be the largest eigenvalue of $\Delta_\Phi$.
    Let \cref{eq:model-in,eq:model-out,eq:iter} be the input, the output, and the implicit graph diffusion layers of DIGNN-$\Delta_\Phi$ respectively.
    Suppose that \cref{assump:loss,assump:output} hold and the node embeddings $\widetilde{\mX}$ of the first layer (\cref{eq:model-in}) of DIGNN-$\Delta_\Phi$ satisfy \cref{assump:embedding}. 
    Then, for any integer $T \geq 1$, any $\delta > 0$, with probability of at least $1-\delta$ over the choice of the training set from $\gG$, for all DIGNN-$\Delta_\Phi$ model $f \in \gH$,
    \begin{equation*}
        L_u(f) \leq \widehat{L}_m(f) + \rho_\ell\rho_h\frac{c_X}{\sqrt{n}}\sum_{t=0}^{T-1}\left(\frac{\lambda_\Phi}{\mu}\right)^{t} + c_0c_1\sqrt{\min(m, u)} + \sqrt{\frac{c_1c_2}{2}\ln\frac{1}{\delta}},
    \end{equation*}
    where $c_0 := \sqrt{32\ln(4e) / 3} < 5.05, c_1 := (\frac{1}{m} + \frac{1}{u})$, and $c_2 := \frac{m+u}{(m+u-1/2)(1-1/2(\max(m, u)))}$.
    Suppose further that \cref{assump:graph,assump:param} hold, we have
    \begin{equation*}
        L_u(f) \leq \widehat{L}_m(f) +  \rho_\ell\rho_h\frac{c_X}{\sqrt{n}}\sum_{t=0}^{T-1}\left(\frac{2c_\phi^2c_\chi c_X\cosh(c_\chi c_X)}{\mu}\right)^{t} + c_0c_1\sqrt{\min(m, u)} + \sqrt{\frac{c_1c_2}{2}\ln\frac{1}{\delta}}.
    \end{equation*}
}
\begin{proof}
    By \cref{app:thrm:transductive,app:thrm:transductive2rade,app:thrm:dignn-rade}, we obtain
    \begin{align*}
        L_u(f) \leq {} & \widehat{L}_m(f) + \widetilde{\Re}_{m+u}(\ell \circ h_{\Theta^{(2)}} \circ \gF_T) + \sqrt{\frac{c_1c_2}{2}\ln\frac{1}{\delta}} \tag{\cref{app:thrm:transductive}} \\
        \leq {} & \widehat{L}_m(f) + \Re_n(\ell \circ h_{\Theta^{(2)}} \circ \gF_T) + \sqrt{\frac{c_1c_2}{2}\ln\frac{1}{\delta}} \tag{\cref{app:thrm:transductive2rade}} \\
        \leq {} & \widehat{L}_m(f) + \rho_\ell\rho_h\frac{c_X}{\sqrt{n}}\sum_{t=0}^{T-1}\left(\frac{\lambda_\Phi}{\mu}\right)^{t} + \sqrt{\frac{c_1c_2}{2}\ln\frac{1}{\delta}}. \tag{\cref{app:thrm:dignn-rade}}
    \end{align*}

    Furthermore, if \cref{assump:graph,assump:param} hold, again by \cref{app:thrm:dignn-rade}, we have
    \begin{equation*}
        L_u(f) \leq \widehat{L}_m(f) +  \rho_\ell\rho_h\frac{c_X}{\sqrt{n}}\sum_{t=0}^{T-1}\left(\frac{2c_\phi^2c_\chi c_X\cosh(c_\chi c_X)}{\mu}\right)^{t} + c_0c_1\sqrt{\min(m, u)} + \sqrt{\frac{c_1c_2}{2}\ln\frac{1}{\delta}}.
    \end{equation*}
\end{proof}

\subsubsection{Proof of \cref{thrm:dignn-main}}

\textbf{\cref{thrm:dignn-main}} (Convergence and Generalization of DIGNN-$\Delta_\Phi$).\textit{
    Under the same settings of \cref{thrm:dignn-generalization}, denote by $\mZ^\star$ the optimal solution of $\mZ = \widetilde{\mX} - \frac{1}{\mu}\Delta_\Phi\mZ$. 
    Suppose that \cref{assump:graph,assump:embedding,assump:param,assump:loss,assump:output} hold and $\|\mZ^\star\| \leq c_Z$. 
    Let $\mu > 2c_\phi^2c_\chi c_X\cosh(c_\chi c_X)$. For any $c_Z > \epsilon > 0, \delta > 0$, let $T = \frac{\log c_Z - \log\epsilon}{-\log\left(2c_\phi^2c_\chi c_X\cosh(c_\chi c_X) / \mu\right)}$, then $\|\mZ^{(T)} - \mZ^\star\| \leq \epsilon$ and \cref{eq:dignn-gen2} holds with probability at least $1-\delta$ over the choice of training set from $\gG$. Moreover, for $T \rightarrow \infty$, we have $\lim_{T \rightarrow \infty}\mZ^{(T)} = \mZ^\star$ and with probability at least $1-\delta$, for all DIGNN-$\Delta_\Phi$ model $f \in \gH$,
    \begin{equation*}
        L_u(f) \leq \widehat{L}_m(f) +  \frac{\rho_\ell\rho_hc_X}{\sqrt{n}(1 - 2c_\phi^2c_\chi c_X\cosh(c_\chi c_X) / \mu)} + c_0c_1\sqrt{\min(m, u)} + \sqrt{\frac{c_1c_2}{2}\log\frac{1}{\delta}}.
    \end{equation*}
}
\begin{proof}
    By $c_Z > \epsilon > 0, \delta > 0, \mu > $ and $T = \frac{\log c_Z - \log\epsilon}{-\log\left(2c_\phi^2c_\chi c_X\cosh(c_\chi c_X) / \mu\right)}$, we have
    \begin{equation*}
        c_Z\left(\frac{2c_\phi^2c_\chi c_X\cosh(c_\chi c_X)}{\mu}\right)^T \leq \epsilon.
    \end{equation*}

    Then, by \cref{thrm:corol3} and \cref{thrm:thrm1}, we obtain
    \begin{equation*}
        \|\mZ^{(T)} - \mZ^\star\| \leq c_Z\left(\frac{\lambda_\Phi}{\mu}\right)^T \leq c_Z\left(\frac{2c_\phi^2c_\chi c_X\cosh(c_\chi c_X)}{\mu}\right)^T \leq \epsilon.
    \end{equation*}
    
    Additionally, \cref{eq:dignn-gen2} holds with $T = \frac{\log c_Z - \log\epsilon}{-\log\left(2c_\phi^2c_\chi c_X\cosh(c_\chi c_X) / \mu\right)}$ by \cref{thrm:dignn-generalization}.

    If $T \rightarrow \infty$, again by \cref{thrm:corol3}, we have
    \begin{equation*}
        \lim_{T \rightarrow \infty}\|\mZ^{(T)} - \mZ^\star\| \leq \lim_{T \rightarrow \infty}c_Z\left(\frac{2c_\phi^2c_\chi c_X\cosh(c_\chi c_X)}{\mu}\right)^T = 0,
    \end{equation*}
    which indicates that $\lim_{T \rightarrow \infty}\mZ^{(T)} = \mZ^\star$. Moreover, by
    \begin{equation*}
        \lim_{T \rightarrow \infty}\sum_{t=0}^{T-1}\left(\frac{2c_\phi^2c_\chi c_X\cosh(c_\chi c_X)}{\mu}\right)^{t} = \lim_{T \rightarrow \infty}\left(\frac{1 - (2c_\phi^2c_\chi c_X\cosh(c_\chi c_X) / \mu)^T}{1 - 2c_\phi^2c_\chi c_X\cosh(c_\chi c_X) / \mu}\right) = \frac{1}{1 - 2c_\phi^2c_\chi c_X\cosh(c_\chi c_X) / \mu},
    \end{equation*}
    we obtain \cref{eq:dignn-gen3}.
\end{proof}

\clearpage
\section{Omitted Derivations}

\subsection{Derivation of parametrized Dirichlet energy}\label{app:der:Dirichlet}
\begin{align*}
    \gS(f) = {} & \|\nabla f\|_\gE^2 \notag \\
    = {} & \langle \nabla f, \nabla f \rangle_\gE \notag \\
    = {} & \langle f, \Delta f \rangle_\gV \notag \\
    = {} & \sum_{i=1}^N\left\langle f(i), (\Delta f)(i) \right\rangle\chi(i) \notag \\
    = {} & \sum_{i=1}^N\left\langle f(i), \frac{1}{\chi(i)}\sum_{j=1}^N\varphi([i, j])^2\phi([i, j])\left(f(i) - f(j)\right) \right\rangle\chi(i) \notag \\
    = {} & \sum_{i=1}^N\sum_{j=1}^N\varphi([i, j])^2\phi([i, j])\left\langle f(i), f(i) - f(j) \right\rangle \notag \\
    = {} & \frac{1}{2}\left(\sum_{i=1}^N\sum_{j=1}^N\varphi([i, j])^2\phi([i, j])\langle f(i), f(i) - f(j) \rangle +  \sum_{i=1}^N\sum_{j=1}^N\varphi([i, j])^2\phi([i, j])\langle f(i), f(i) - f(j) \rangle\right) \notag \\
    = {} & \frac{1}{2}\left(\sum_{j=1}^N\sum_{i=1}^N\varphi([j, i])^2\phi([j, i])\langle f(j), f(j) - f(i) \rangle + \sum_{i=1}^N\sum_{j=1}^N\varphi([i, j])^2\phi([i, j])f(i)\left(f(i) - f(j)\right)\right) \tag{exchange $i$ and $j$ for the first term} \notag \\
    = {} & \frac{1}{2}\left(\sum_{i=1}^N\sum_{j=1}^N\varphi([i, j])^2\phi([i, j])\langle f(j), f(j) - f(i) \rangle - \sum_{i=1}^N\sum_{j=1}^N\varphi([i, j])^2\phi([i, j])\langle f(i), f(j) - f(i) \rangle\right) \tag{$[j, i] = [i, j]$} \notag \\
    = {} & \frac{1}{2}\sum_{i=1}^N\sum_{j=1}^N\varphi([i, j])^2\phi([i, j])\langle f(j) - f(i), f(j) - f(i) \rangle \notag \\
    = {} & \frac{1}{2}\sum_{i=1}^N\sum_{j=1}^N\varphi([i, j])^2\phi([i, j])\|f(j) - f(i)\|^2.
\end{align*}

\subsection{Derivation of the fixed-point equilibrium equation}\label{app:der:equi}
Recall the objective function:
\begin{equation*}
    \inf_f\gL(f) = \inf_f\left[\frac{1}{2}\sum_{i=1}^N\sum_{j=1}^N\varphi([i, j])^2\phi([i, j])\|f(j) - f(i)\|^2 + \mu\sum_{i \in \gI}\|f(i) - \rvy(i)\|_\gV^2\right].
\end{equation*}
By \cref{thrm:lem2} we have
\begin{equation*}
    \left.\frac{\od\gS(f)}{\od f}\right|_i = 2\chi(i)(\Delta f)(i).
\end{equation*}
Note also that for $i \in \gI$,
\begin{align}
    \left.\frac{\od\sum_{i \in \gI}\|f(i) - \rvy(i)\|_\gV^2}{\od f}\right|_i = {} & \left.\frac{\od\sum_{i \in \gI}\langle f(i) - \rvy(i), f(i) - \rvy(i) \rangle_\gV}{\od f}\right|_i \notag \\
    = {} & \left.\frac{\od\sum_{i \in \gI}\|f(i) - \rvy(i)\|^2\chi(i)}{\od f}\right|_i \notag \\
    = {} & 2(f(i) - \rvy(i))\chi(i).
\end{align}

For labeled nodes $i \in \gI$, we have
\begin{align}
    \left.\frac{\od\gL(f)}{\od f}\right|_i = {} & \left.\frac{\od\gS(f)}{\od f}\right|_i + 2\mu\chi(i)(f(i) - \rvy(i)) \notag \\
    = {} & 2\chi(i)(\Delta f)(i) + 2\mu\chi(i)(f(i) - \rvy(i)).
\end{align}

\begin{align}
    \left.\frac{\od\gL(f)}{\od f}\right|_i = 0 \Longrightarrow {} & 2\chi(i)(\Delta f^\star)(i) + 2\mu\chi(i)(f^\star(i) - \rvy(i)) = 0 \notag \\
    \Longrightarrow {} & f^\star(i) = \rvy(i) - \frac{1}{\mu}(\Delta f^\star)(i).
\end{align}

For unlabeled nodes $i \notin \gI$,
\begin{equation}
    \left.\frac{\od\gL(f)}{\od f}\right|_i = 
 \left.\frac{\od\gS(f)}{\od f}\right|_i = 2\chi(i)(\Delta f)(i).
\end{equation}

\begin{align}
    \left.\frac{\od\gL(f)}{\od f}\right|_i = 0 \Longrightarrow 2\chi(i)(\Delta f^\star)(i) = {} & 2\sum_{j=1}^N\varphi([i, j])^2\phi([i, j])\left(f^\star(i) - f^\star(j)\right) \notag \\
    = {} & 2\sum_{j=1}^N\varphi([i, j])^2\phi([i, j])f^\star(i) - 2\sum_{j=1}^N\varphi([i, j])^2\phi([i, j])f^\star(j) \notag \\
    = {} & 0.
\end{align}
\begin{align}
    \Longrightarrow {} & \sum_{j=1}^N\varphi([i, j])^2\phi([i, j])f^\star(i) = \sum_{j=1}^N\varphi([i, j])^2\phi([i, j])f^\star(j) \notag \\
    \Longrightarrow {} & f^\star(i) = \sum_{j=1}^N\frac{\varphi([i, j])^2\phi([i, j])}{\sum_{k=1}^N\varphi([i, k])^2\phi([i, k])}f^\star(j) = (\mP f^\star)(i)
\end{align}

Because for all $i \in \gV$, 
\begin{equation}
    \mC_{i,:} = (1 - \delta_i)\mP_{i,:} - \delta_i\frac{1}{\mu}\Delta_{i,:} =
    \begin{cases}
        -\frac{1}{\mu}\Delta_{i,:} & \text{if } i \in \gI, \\
        \mP_{i,:} & \text{otherwise}.
    \end{cases}
\end{equation}
we have $f(i) = \rvy'(i) + (\mC f)(i)$ for all $i \in \gV$, i.e.,
\begin{equation}
    f = \mY' + \mC f.
\end{equation}

\clearpage

\section{Training For DIGNNs}\label{app:train-dignn}

\subsection{Forward pass}
With appropriately chosen hyperparameter $\mu > \lambda_{\text{max}}$, \cref{thrm:corol3} illustrates that the equilibrium of the fixed-point equation~\cref{eq:diff} can be obtained by iterating \cref{eq:iter}. 
Therefore, for the forward evaluation, we can simply iterate \cref{eq:iter}.

\subsection{Backward pass}
For the backward pass, we can use implicit differentiation~\citep{ImplicitBook-2002,DEQ-NIPS2019} to compute the gradients of trainable parameters by directly differentiating through the equilibrium.
We first introduce the matrix vectorization and the Kronecker product and then provide the derivation of the gradients of trainable parameters with implicit differentiation. 

\subsubsection{Matrix vectorization}
For a matrix $\mA \in \R^{m \times n}$, we define its vectorization $\vect(\mA)$ as
\begin{equation*}
    \vect(\mA) = (\emA_{1,1}, \dots, \emA_{m,1}, \emA_{1,2}, \dots, \emA_{m,2}, \dots, \emA_{1,n}, \dots, \emA_{m,n})^\top.
\end{equation*}

\subsubsection{Kronecker product}
For two matrices $\mA \in \R^{m \times n}$ and $\mB \in \R^{p \times q}$, the Kronecker product $\mA \otimes \mB \in \R^{pm \times qn}$ is defined as
\begin{equation*}
    \mA \otimes \mB = \left[
    \begin{array}{ccc}
        \emA_{1,1}\mB & \dots & \emA_{1,n}\mB \\
        \vdots & \ddots & \vdots \\
        \emA_{m,1}\mB & \dots & \emA_{m,n}\mB
    \end{array}
    \right].
\end{equation*}
By the definition of the Kronecker product, we have the following property of the vectorization with the Kronecker product~\citep{Kron-Year2013}:
\begin{equation*}
    \vect(\mA\mB) = (\mI_m \otimes \mA)\vect(\mB).
\end{equation*}

\subsubsection{Backward pass gradient computation by implicit differentiation}
Denote the loss function by
\begin{equation*}
    \ell = \ell(\mY, \widehat{\mY}) = \ell(\mY, h_{\Theta^{(2)}}(\mZ^\star)).
\end{equation*}
Writing the fixed-point equation~\cref{eq:diff} into the vectorization form:
\begin{align}
    \vect(\mZ^\star) = {} & \vect(\mX - \frac{1}{\mu}\Delta\mZ^\star) \notag \\
    = {} & \vect(\mX) - \frac{1}{\mu}\vect(\Delta\mZ^\star) \notag \\
    = {} & \vect(\mX) - \frac{1}{\mu}(\mI \otimes \Delta)\vect(\mZ^\star). \label{app:eq:vec_diff}
\end{align}
For simplicity, denote by $\vec{\mX} := \vect(\mX)$ and $\vec{\mZ}^\star := \vect(\mZ^\star)$. Then \cref{app:eq:vec_diff} becomes
\begin{equation*}
    \vec{\mZ}^\star = \vec{\mX} - \frac{1}{\mu}(\mI \otimes \Delta)\vec{\mZ}^\star.
\end{equation*}
By the chain rule, we have
\begin{equation*}
    \frac{\partial\ell}{\partial(\cdot)} = \frac{\partial\ell}{\partial\vec{\mZ}^\star}\frac{\od\vec{\mZ}^\star}{\od(\cdot)} = \frac{\partial\ell}{\partial h_\Theta}\frac{\partial h_\Theta}{\partial\vec{\mZ}^\star}\frac{\od\vec{\mZ}^\star}{\od(\cdot)}
\end{equation*}

Since $\vec{\mZ}^*$ and $(\cdot)$ are implicitly related, $\frac{\od\vec{\mZ}^\star}{\od(\cdot)}$ cannot be directly obtained by automatic differentiation packages. By differentiating both sides of the equilibrium equation $\vec{\mZ}^\star = \vec{\mX} - \frac{1}{\mu}(\mI \otimes \Delta)\vec{\mZ}^\star$, we have
\begin{align*}
    \frac{\od\vec{\mZ}^\star}{\od(\cdot)} = {} & \frac{\od(\vec{\mX} - \frac{1}{\mu}(\mI \otimes \Delta)\vec{\mZ}^\star)}{\od(\cdot)} \\
    = {} & \frac{\partial(\vec{\mX} - \frac{1}{\mu}(\mI \otimes \Delta)\vec{\mZ}^\star)}{\partial(\cdot)} + \frac{\partial(\vec{\mX} - \frac{1}{\mu}(\mI \otimes \Delta)\vec{\mZ}^\star)}{\partial\vec{\mZ}^\star}\frac{\od\vec{\mZ}^\star}{\od(\cdot)} \\
    = {} & \frac{\partial(\vec{\mX} - \frac{1}{\mu}(\mI \otimes \Delta)\vec{\mZ}^\star)}{\partial(\cdot)} - \frac{1}{\mu}\frac{\partial(\mI \otimes \Delta)\vec{\mZ}^\star}{\partial\vec{\mZ}^\star}\frac{\od\vec{\mZ}^\star}{\od(\cdot)}.
\end{align*}

Rearrange the above equation, we have
\begin{equation*}
    \left(\mI + \frac{1}{\mu}\frac{\partial(\mI \otimes \Delta)\vec{\mZ}^\star}{\partial\vec{\mZ}^\star}\right)\frac{\od\vec{\mZ}^\star}{\od(\cdot)} = \frac{\partial(\vec{\mX} - \frac{1}{\mu}(\mI \otimes \Delta)\vec{\mZ}^\star)}{\partial(\cdot)},
\end{equation*}
which implies that
\begin{align*}
    \frac{\od\vec{\mZ}^\star}{\od(\cdot)} = {} & \left(\mI + \frac{1}{\mu}\frac{\partial(\mI \otimes \Delta)\vec{\mZ}^\star}{\partial\vec{\mZ}^\star}\right)^{-1}\frac{\partial(\vec{\mX} - \frac{1}{\mu}(\mI \otimes \Delta)\vec{\mZ}^\star)}{\partial(\cdot)} \notag \\
    = {} & -\frac{1}{\mu}\left(\mI + \frac{1}{\mu}\frac{\partial(\mI \otimes \Delta)\vec{\mZ}^\star}{\partial\vec{\mZ}^\star}\right)^{-1}\frac{\partial(\mI \otimes \Delta)\vec{\mZ}^\star}{\partial(\cdot)}.
\end{align*}

Let
\begin{equation*}
    \left.J\right|_{\vec{\mZ}^\star} = \mu\left(\mI + \frac{1}{\mu}\frac{\partial(\mI \otimes \Delta)\vec{\mZ}^\star}{\partial\vec{\mZ}^\star}\right) = \left(\mu\mI + \frac{\partial(\mI \otimes \Delta)\vec{\mZ}^\star}{\partial\vec{\mZ}^\star}\right),
\end{equation*}
we obtain
\begin{align*}
    \frac{\partial\ell}{\partial(\cdot)} = {} & \frac{\partial\ell}{\partial\vec{\mZ}^\star}\frac{\od\vec{\mZ}^\star}{\od(\cdot)} \\
    = {} & -\frac{\partial\ell}{\partial\vec{\mZ}^\star}\left(\left.J^{-1}\right|_{\vec{\mZ}^\star}\right)\frac{\partial(\mI \otimes \Delta)\vec{\mZ}^\star}{\partial(\cdot)} \\
    = {} & -\frac{\partial\ell}{\partial h_{\Theta^{(2)}}}\frac{\partial h_{\Theta^{(2)}}}{\partial\vec{\mZ}^\star}\left(\left.J^{-1}\right|_{\vec{\mZ}^\star}\right)\frac{\partial(\mI \otimes \Delta)\vec{\mZ}^\star}{\partial(\cdot)}.
\end{align*}
Note that the computation cost of the inverse Jacobian $J^{-1}$ is expensive. To compute $-\frac{\partial\ell}{\partial\vec{\mZ}^\star}\left(\left.J^{-1}\right|_{\vec{\mZ}^\star}\right)$, we can alternatively solve the linear system, which is more efficient to compute:
\begin{equation}
    \left(\left.J^\top\right|_{\vec{\mZ}^\star}\right)\vu^\top + \left(\frac{\partial\ell}{\partial\vec{\mZ}^\star}\right)^\top = \mathbf{0},
\end{equation}
where the vector-Jacobian product can be efficiently obtained by automatic differentiation packages (e.g., PyTorch) for any $\vu$ without explicitly writing out the Jacobian matrix.

\clearpage
\section{Canonical Graph Laplacian Operators and Dirichlet Energies}\label{sec:app:canonical}
\subsection{Canonical graph Laplcian operators}\label{app:lap}
Here we discuss the connection between our definition of graph Laplacian operator given in \cref{def:graph-lap} with canonical graph Laplacian operators, including unnormalized graph Laplacian, random walk graph Laplacian, and normalized graph Laplacian~\citep{GraphTheory-Book1997,Lap-JMLR2007}. \cref{tab:tab1} summarizes the comparisons of Hilbert spaces and positive real-valued functions between graph neural Laplacian, unnormalized graph Laplacian, random walk graph Laplacian, and normalized graph Laplacian.

\begin{table}[htp]
    \centering
    \caption{Graph Laplacian Operators.}\label{tab:tab1}
    \resizebox{\linewidth}{!}%
    {
    \begin{tabular}{lll}
        \toprule
         Type & Hilbert spaces & Positive real-valued function \\
        \midrule
        Unnormalized $\Delta^{\text{(un)}}$ & Fixed & $\chi(i) = 1, \varphi^2([i, j])\phi([i, j]) = \emA_{i,j}$ \\
        Random walk $\Delta^{\text{(rw)}}$ & Fixed & $\chi(i) = \emD_i, \varphi^2([i, j])\phi([i, j]) = \emA_{i,j}$ \\
        Normalized\tablefootnote{The graph gradient for the normalized case is defined as $(\nabla f)([i, j]) = \varphi([i, j])(f(j)/\sqrt{\emD_j} - f(i)/\sqrt{\emD_i})$.} $\Delta^{\text{(n)}}$ & Fixed & $\chi(i) = \sqrt{\emD_i}, \phi^2([i, j])\phi([i, j]) = \emA_{i,j}$ \\
        \midrule
        \multirow{3}{*}{Parameterized $\Delta_\Phi$ } & \multirow{3}{*}{Learnable} & $\chi(i) = \emD_i\tanh(\|\Theta_\chi\vx_i\|)$ \\ 
        & & $\phi([i, j]) = \tanh(|(\Theta_\phi\Theta_\chi\vx_i)^\top\Theta_\phi\Theta_\chi\vx_j|)$ \\ 
        & & $\varphi([i, j]) = \sqrt{\emA_{i,j}\tanh((\|\Theta_\varphi(\vx_i - \vx_j)\| + \epsilon)^{-1})}$ \\
        \bottomrule
    \end{tabular}
    }
\end{table}

\textbf{Unnormalized graph Laplacian operator.} 
Given a graph $\gG = (\gV, \gE, \mA)$ and a function $f: \gV \mapsto \R$, the unormalized graph Laplacian operator $\Delta^{\text{(un)}}$ is given by:
\begin{equation}\label{app:eq:un-lap}
    (\Delta^{\text{(un)}}f)(i) = \sum_{j=1}^N\emA_{i,j}\left(f(i) - f(j)\right), \quad \text{for all } i \in \gV.
\end{equation}
The unnormalized graph Laplacian operator corresponds to our graph Laplacian operator when we adopt the following fixed positive real-value functions $\chi, \phi$ and $\varphi$:
\begin{equation}\label{app:eq:un-func}
    \begin{aligned}
    \chi(i) = {} & 1, \quad \text{for all } i \in \gV, \\
    \varphi^2([i, j])\phi([i, j]) = {} & \emA_{i,j}, \quad \text{for all } [i, j] \in \gE,
    \end{aligned}
\end{equation}
which can be easily verified by substituting the above $\chi, \phi, \varphi$ into \cref{eq:param-lap} to obtain \cref{app:eq:un-lap}.

\textbf{Random walk graph Laplacian operator.} 
Given a graph $\gG = (\gV, \gE, \mA)$ and a function $f: \gV \mapsto \R$, the random walk graph Laplacian operator $\Delta^{\text{(rw)}}$ is given by:
\begin{equation}\label{app:eq:rw-lap}
    (\Delta^{\text{(rw)}}f)(i) = \sum_{j=1}^N\frac{\emA_{i,j}}{\emD_i}\left(f(i) - f(j)\right), \quad \text{for all } i \in \gV.
\end{equation}
Similarly, the random walk graph Laplacian operator corresponds to our graph Laplacian operator when we adopt the following fixed $\chi, \phi$ and $\varphi$:
\begin{equation}\label{app:eq:rw-func}
    \begin{aligned}
    \chi(i) = {} & \emD_i, \quad \text{for all } i \in \gV, \\
    \varphi^2([i, j])\phi([i, j]) = {} & \emA_{i,j}, \quad \text{for all } [i, j] \in \gE,
    \end{aligned}
\end{equation}
which can be easily verified by substituting $\chi, \phi, \varphi$ in \cref{app:eq:rw-func} into \cref{eq:param-lap} to obtain \cref{app:eq:rw-lap}.

\textbf{Normalized graph Laplacian operator~\citep{GraphTheory-Book1997,SSL-NIPS2004}.} 
Following the same inner product on vertex space and edge space defined in \cref{def:def2,def:def3} respectively, the form of the graph gradient operator used in the normalized graph Laplacian operator is different from our graph gradient operator defined in \cref{def:def4}.
\begin{definition}[Normalized Graph Gradient]
    Given a graph $\gG = (\gV, \gE)$ and a function $f: \gV \mapsto \R$, the normalized graph gradient $\nabla^{\text{(n)}}: \gH(\gV, \chi) \mapsto \gH(\gE, \phi)$ is defined as follows:
    \begin{equation}
        (\nabla^{\text{(n)}} f)([i, j]) = \varphi([i, j])\left(\frac{f(j)}{\sqrt{\emD_j}} - \frac{f(i)}{\sqrt{\emD_i}}\right), \text{ for all } [i, j] \in \gE,
    \end{equation}
\end{definition}
The normalized graph divergence operator $\divg^{\text{(n)}}: \gH(\gE, \phi) \mapsto \gH(\gV, \chi)$ can be derived by satisfying $\langle \nabla^{\text{(n)}} f, g \rangle_\gE = \langle f, -\divg^{\text{(n)}}g \rangle_\gV$ for any functions $f: \gV \mapsto \R$ ad $g: \gE \mapsto \R$. Based on the normalized graph gradient operator and normalized graph divergence operator, the normalized graph Laplacian operator $\Delta^{\text{(n)}}$ is given by~\citep{pGNN-ICML2022}:
\begin{equation}\label{app:eq:n-lap}
    (\Delta^{\text{(n)}}f)(i) = \sum_{j=1}^N\frac{\emA_{i,j}}{\sqrt{\emD_i}}\left(\frac{f(i)}{\sqrt{\emD_i}} - \frac{f(j)}{\sqrt{\emD_j}}\right), \quad \text{for all } i \in \gV.
\end{equation}
Then, the normalized graph Laplacian operator corresponds to the case when we adopt the following fixed $\chi, \phi$ and $\varphi$:
\begin{equation}
    \begin{aligned}
    \chi(i) = {} & \sqrt{\emD_i} \quad \text{for all } i \in \gV, \\
    \varphi^2([i, j])\phi([i, j]) = {} & \emA_{i,j}, \quad \text{ for all } [i, j] \in \gE.
    \end{aligned}
\end{equation}
We omit the detail derivation here, which should be easy to verify following a similar way to derive the explicit expression of the graph Laplacian operator given in \cref{thrm:lem2}.
\begin{remark}[The Hilbert spaces for canonical graph Laplacian operators]
    It is worth noting that choosing the above canonical graph Laplacian operators $\Delta^{\text{(un)}}, \Delta^{\text{(rw)}}$, or $\Delta^{\text{(n)}}$, the functions $\phi$ and $\varphi$ are not fixed. Therefore, the corresponding Hilbert space on edges and the graph gradient operator are not explicitly specified, which could lead to confusion in modeling the graph metrics. As a result, graph learning methods that are developed based on these canonical Laplacian operators, for examples GNNs~\citep{GCN-ICLR2017,APPNP-ICLR2019,IGNN-NIPS2020}, graph-based semi-supervised learning algorithms~\cite{SSL-NIPS2004,SSL-COLT2003,SSL-ICML2003}, may have limited adaptability to learn graph metrics and potentially hinder their performance in graph learning problems. On the contrary, in the definition of our graph Laplacian operator, we use $\chi, \phi, \varphi$ to explicitly learn the Hilbert space on vertices, the Hilbert space on edges, and the graph gradient operator, respectively. Therefore, it may have more flexibility to model the graph metrics. 
\end{remark}

\subsection{Canonical Dirichlet energies}
Here we introduce the Dirichlet energies induced by canonical graph Laplacian operators.

\textbf{Dirichlet energy induced by unnormalized and random walk graph Laplacians operators.} 
For the unnormalized and random walk graph Laplacian operators $\Delta^{\text{(un)}}, \Delta^{\text{(rw)}}$, $\varphi^2([i, j])\phi([i, j]) = \emA_{i,j}$ for all $[i, j] \in \gE$. Then by \cref{eq:Dirichlet}, the Dirichlet energy $\gS^{\text{(un)}}$ (or $\gS^{\text{(rw)}}$) induced by $\Delta^{\text{(un)}}$ (or $\Delta^{\text{(rw)}}$) is given by
\begin{align}
    \gS^{\text{(un)}}(f) = \gS^{\text{(rw)}}(f) = {} &  \frac{1}{2}\sum_{i=1}^N\sum_{j=1}^N\varphi([i, j])^2\phi([i, j])\|f(j) - f(i)\|^2 \notag \\
    = {} & \frac{1}{2}\sum_{i=1}^N\sum_{j=1}^N\emA_{i,j}\|f(j) - f(i)\|^2.
\end{align}
By \cref{thrm:lem3}, we have
\begin{align}
    \left.\frac{\od\gS^{\text{(un)}}(f)}{\od f}\right|_i = \left.\frac{\od\gS^{\text{(rw)}}(f)}{\od f}\right|_i = {} & 2\sum_{j=1}^N\varphi([i, j])^2\phi([i, j])(f(i) - f(j)) \notag \\
    = {} & 2\sum_{j=1}^N\emA_{i,j}(f(i) - f(j)) \notag \\
    = {} & 2\Delta^{\text{(un)}}f(i) = 2\emD_i\Delta^{\text{(rw)}}f(i).
\end{align}

\paragraph{Dirichlet energy induced by the normalized graph Laplacian operator.}
For the normalized graph Laplacian operator $\Delta^{\text{(n)}}$, $\varphi^2([i, j])\phi([i, j]) = \emA_{i,j}$ for all $[i, j] \in \gE$. The Dirichlet energy $\gS^{\text{(n)}}$ induced by $\Delta^{\text{(n)}}$ is given by
\begin{align}
    \gS^{\text{(n)}}(f) = {} & \|\nabla^{\text{(n)}}f\|_\gE^2 \notag \\
    = {} & \frac{1}{2}\sum_{i=1}^N\sum_{j=1}^N\varphi([i, j])^2\phi([i, j])\left\|\frac{f(j)}{\sqrt{\emD_j}} - \frac{f(i)}{\sqrt{\emD_i}}\right\|^2 \notag \\
    = {} & \frac{1}{2}\sum_{i=1}^N\sum_{j=1}^N\emA_{i,j}\left\|\frac{f(j)}{\sqrt{\emD_j}} - \frac{f(i)}{\sqrt{\emD_i}}\right\|^2.
\end{align}
Then we have
\begin{align}
    \left.\frac{\od\gS^{\text{(n)}}(f)}{\od f}\right|_i = {} & 2\sum_{j=1}^N\frac{\emA_{i,j}}{\sqrt{\emD_i}}\left(\frac{f(i)}{\sqrt{\emD_i}} - \frac{f(j)}{\sqrt{\emD_j}}\right) \notag \\
    = {} & 2\Delta^{\text{(n)}}f(i).
\end{align}

\clearpage
\section{Experimental Setup and Additional Experiments}\label{app:exp}

\subsection{Experimental setup}

\subsubsection{Dataset description}
\paragraph{Semi-supervised node classification.}
For semi-supervised node classification, we use both homophilic and heterophilic datasets to demonstrate the effectiveness of our framework. 
The homophilic datasets are the citation networks including Cora, CiteSeer, and PubMed. 
Regarding the heterophilic datasets, we avoid using very small datasets due to the potential sensitivity of GNN performance to different random splitting settings and the risk of overfitting. 
Therefore, we utilize larger heterophilic datasets such as Chameleon, Squirrel, Penn94, Cornell5, and Amherst41, each containing at least $1000$ nodes in the graph. 
More information on these new heterophilic datasets can be found in \citep{LINKX-NIPS2021}. Dataset statistics are provided in the Caption of \cref{tab:tab2,tab:tab5}.

\paragraph{Graph classification.}
For the graph classification task, we utilize well-known datasets, including MUTAG, PTC, PROTEINS, NCI1, IMDB-BINARY, and IMDB-MULTI, which are available in the TUDataset benchmark suite~\citep{morris2020tudataset}.
MUTAG, PTC, PROTEINS, and NCI1 are collected from the bioinformatics domain while IMDB-BINARY and IMDB-MULTI are social network datasets. 
Specifically, graphs in MUTAG represent mutagenic aromatic and heteroaromatic compounds. 
PTC consists of compounds with carcinogenicity for males and females. 
NCI1 dataset consists of chemical compounds with the ability to suppress the growth of tumor cells made available by the National Cancer Institute (NCI). 
PROTEINS dataset consists of graphs with nodes as secondary structure elements (SSEs) with adjacent nodes indicating the neighborhood of the amino-acid sequence or 3D space.
For all the graph classification datasets, we follow the 10-fold splits from \cite{GIND-ICML2022} for training, validation, and testing.

\subsubsection{Model architectures}
For graph classification tasks and node classification tasks on homophilic graphs, our model architecture consists of a 1-layer MLP layer and an implicit graph diffusion layer followed by an output classifier. 
For node classification tasks on heterophilic graphs, our model architecture consists of an input feature preprocessing unit layer and an implicit graph diffusion layer followed by an output classifier. Specifically,, we use $h_{\Theta^{(1)}}(\mA, \mX) =  \mA \mX$ as preprocessing step for Chameleon and Squirrel datasets. For Penn94, Cornell91 and Amherst41 datasets, we use the LINKX structure of $ \textrm{MLP} ( \text{Concat} ( \textrm{MLP} (\mA), \textrm{MLP} (\mX)))$ as a preprocessing step. For a fair comparison, we add the same LINKX preprocessing step to other Implicit GNN models and report the best scores achieved, with and without the preprocessing step. We use dropout after each layer and batch normalization before the Implicit layer. We implement our code in Pytorch~\citep{paszke2017automatic} along with PyTorch-geometric libraries~\citep{pyg-aXiv2019}.
For heterophilic datasets of Penn94, Cornell41, and Amherst41, we use the source codes of other baselines and report the best results obtained after a substantial hyperparameter search.

\subsubsection{Hyperparameters}

We search the hyperparameters for our model from the following.
\begin{itemize}
    \item hidden dimension $\in \{16,32, 64, 128\}$
    \item $\mu \in \{1, 1.25, 2, 2.1, 2.2, 2.4, 2.5, 5, 10\}$. 
    \item learning rate $\in \{0.001, 0.005, 0.01\}$
    \item dropout rate $\in \{0, 0.1, 0.25, 0.5, 0.75\}$
    \item maximum iteration of the implicit layer $\in \{2, 4, 10, 20, 100\}$
    \item weight decay $\in \{0, 1e\text{-}4, 1e\text{-}5\}$
\end{itemize}
The hyperparameters used in our experiments are shown in Tables~\ref{tab:tab6} and~\ref{tab:tab7}.

\begin{table}[htp]
    \centering
    \caption{Hyperparameters of DIGNN-$\Delta^{\text{(rw)}}$.}\label{tab:tab6}
    \begin{tabular}{lccccccc}
        \toprule
        Dataset & $\mu$ & num hid & lr & weight decay & max iter & threshold & dropout \\
        \midrule
        Chameleon & 2.20 & 128 & 0.01 & 0 & 10 & 1e-6 & 0.5 \\
        Squirrel & 2.20 & 128 & 0.01 & 0 & 10 & 1e-6 & 0.1 \\
        Penne94 & 2.10 & 32 & 0.001 & 0 & 10 & 1e-6 & 0.75 \\
        Cornell5 & 2.50 & 32 & 0.001 & 0 & 4 & 1e-6 & 0.75 \\
        Amherst41 & 2.50 & 64 & 0.01 & 1e-4 & 20 & 1e-6  & 0.5 \\
        \midrule
        Cora & 2.10 & 64 & 0.001 & 1e-5 & 20 & 1e-6 & 0.75 \\
        CiteSeer & 2.10 & 64 & 0.001 & 0 & 10 & 1e-6 & 0.5 \\
        PubMed & 2.50 & 64 & 0.001 & 1e-5 & 10 & 1e-6 & 0.5 \\
        \midrule
        MUTAG & 2.40 & 64 & 0.005 & 1e-5 & 20 & 1e-6 & 0 \\
        PTC & 2.50 & 64 & 0.001 & 0 & 10 & 1e-6 & 0 \\
        Proteins & 2.50 & 64 & 0.001 & 0 & 10 & 1e-6 & 0 \\
        NCI1 & 2.40 & 64 & 0.001 & 1e-5 & 4 & 1e-6 & 0 \\
        IMDB-B & 2.50 & 128 & 0.001 & 0 & 10 & 1e-6 & 0 \\
        IMDB-M & 2.50 & 64 & 0.001 & 1e-5 & 20 & 1e-6 & 0\\
        
        \midrule
        PPI & 2.00 & 256 & 0.01  & 0 & 50 & 1e-6 & 0.1 \\
        \bottomrule
    \end{tabular}
\end{table}

\begin{table}[htp]
    \centering
    \caption{Hyperparameters of DIGNN-$\Delta_\Phi$.}\label{tab:tab7}
    \begin{tabular}{lccccccc}
        \toprule
        Dataset & $\mu$ & num hid & lr & weight decay & max iter & threshold & dropout \\
        \midrule
        Chameleon & 2.20 & 128 & 0.01 & 0 & 10 & 1e-6 & 0.5 \\
        Squirrel & 2.20 & 128 & 0.01 & 0 & 10 & 1e-6 & 0.1 \\
        Penne94 & 1.25 & 32 & 0.001 & 0 & 10 & 1e-6 & 0.75 \\
        Cornell5 & 2.50 & 32 & 0.001 & 0 & 4 & 1e-6 & 0.75 \\
        Amherst41 & 2.50 & 64 & 0.01 & 0 & 20 & 1e-6  & 0.5 \\
        \midrule
        Cora & 2.10 & 64 & 0.001 & 1e-5 & 20 & 1e-6 & 0.75 \\
        CiteSeer & 2.10 & 64 & 0.001 & 0 & 10 & 1e-6 & 0.5 \\
        PubMed & 2.50 & 64 & 0.001 & 1e-5 & 10 & 1e-6 & 0.5 \\
        \midrule
        MUTAG & 1.25 & 64 & 0.005 & 1e-5 & 20 & 1e-6 & 0 \\
        PTC & 1.25 & 64 & 0.001 & 0 & 10 & 1e-6 & 0 \\
        Proteins & 1.25 & 64 & 0.001 & 0 & 10 & 1e-6 & 0 \\
        NCI1 & 1.25 & 64 & 0.001 & 1e-5 & 4 & 1e-6 & 0 \\
        IMDB-B & 2.50 & 128 & 0.001 & 0 & 10 & 1e-6 & 0 \\
        IMDB-M & 1.25 & 64 & 0.001 & 1e-5 & 20 & 1e-6 & 0\\
        \midrule
        PPI & 2.00 & 256 & 0.01 & 0 & 50 & 1e-6 & 0.1 \\
        \bottomrule
    \end{tabular}
\end{table}

\subsection{Semi-supervised node classification on homophilic graphs}\label{app:exp-homo}
In this section, we conduct node classification experiments on three commonly used homophilic datasets: Cora, CiteSeer, and PubMed~\citep{NodeHomoData-AIM2008} and PPI dataset~\citep{NodePPIData-Bio2017} for multi-label multi-graph inductive learning. 
For Cora, CiteSeer, and PubMed, we use standard train/validation/test splits as in \citep{Geom-GCN-ICLR2020}. 
The results are presented in \cref{tab:tab3}.
For PPI, we use the standard data splits as in \citep{GraphSage-NIPS2017} and results are reported in \cref{tab:tab4}.
The results of all baselines except for GRAND-l on Cora, CiteSeer, PubMed, and PPI are borrowed from \cite{MGNNI-NIPS2022,GIND-ICML2022,LINKX-NIPS2021}, while the results for GRAND-l were obtained by us using their officially published source code. 

\cref{tab:tab3} shows that our models obtain comparable accuracy to all baselines on homophilic datasets Cora, CiteSeer, and PubMed where long-range dependencies might not significantly enhance the prediction performance. 
The results presented in \cref{tab:tab4} for PPI show that DIGNNs ($\Delta^{\text{(rw)}}$ and $\Delta_\Phi$) achieve Micro-F1 scores of $98.9\%$ and $99.1\%$ respectively. 
Our models surpass most explicit and implicit baselines and are close to the SOTA GCNII. 
It suggests that DIGNNs apply to multi-label multi-graph inductive learning.

\begin{table}[tp]
    \centering
    \caption{Results on homophilic node classification datasets: mean accuracy (\%). Best results outlined in bold.}\label{tab:tab3}
    \begin{tabular}{lccc}
        \toprule
        & Cora & CiteSeer & PubMed \\
        \midrule
        GCN & $85.77$ & $73.68$ & $88.13$ \\
        GAT & $86.37$ & $74.32$ & $87.62$ \\
        APPNP & $87.87$ & $76.53$ & $89.40$ \\
        Geom-GCN & $85.27$ & $\bf{77.99}$ & $\bf{90.05}$ \\
        GCNII & $\bf{88.49}$ & $77.08$ & $89.57$ \\
        H2GCN & $87.87$ & $77.11$ & $89.49$ \\
        \midrule
        IGNN & $85.80$ & $75.24$ & $87.66$ \\
        EIGNN & $85.89$ & $75.31$ & $87.92$ \\
        GRAND-l* & $87.02$ & $75.79 $ &  $87.08$\\
        MGNNI & $83.37$ & $75.57$ & $88.03 $ \\
        GIND & $88.25$ & $76.81$ & $89.22$ \\
        \midrule
        DIGNN-$\Delta^{\text{(rw)}}$ & $85.01$ & $75.22$ & $88.65$ \\
        DIGNN-$\Delta_\Phi$ & $86.68$ & $76.98$ & $88.60$ \\
        \bottomrule
    \end{tabular}
\end{table}
\begin{table}
    \centering
    \caption{Results of multi-label node classification on PPI. Best results outlined in bold.}\label{tab:tab4}
    \begin{tabular}{lc}
        \toprule
        Method & Micro-F1 (\%) \\
        \midrule
        GCN & $59.2$ \\
        GraphSAGE & $78.6$ \\
        GAT & $97.3$ \\
        JKNet & $97.6$ \\
        GCNII & $\bf{99.5}$ \\
        \midrule
        IGNN & $97.0$ \\
        EIGNN & $98.0$ \\
        MGNNI & $98.7$ \\
        GIND & $98.4$ \\
        \midrule
        DIGNN-$\Delta^{\text{(rw)}}$ & $98.9$ \\
        DIGNN-$\Delta_\Phi$ & $99.1$ \\
        \bottomrule
    \end{tabular}
\end{table}

\clearpage

\begin{figure}[htp]
    \centering
    \begin{subfigure}[b]{0.4\textwidth}
        \centering
        \includegraphics[width=\textwidth]{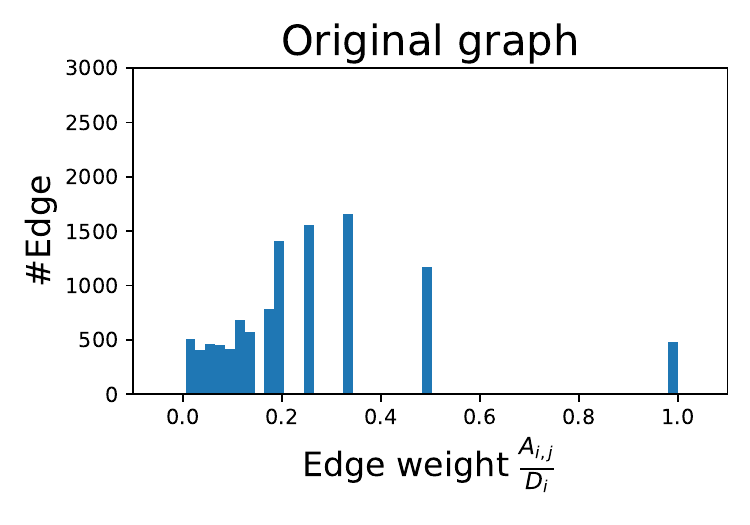}
        \caption{Original edge weight of Cora.}\label{app:fig:cora-org}
    \end{subfigure}
    \begin{subfigure}[b]{0.4\textwidth}
        \centering
        \includegraphics[width=\textwidth]{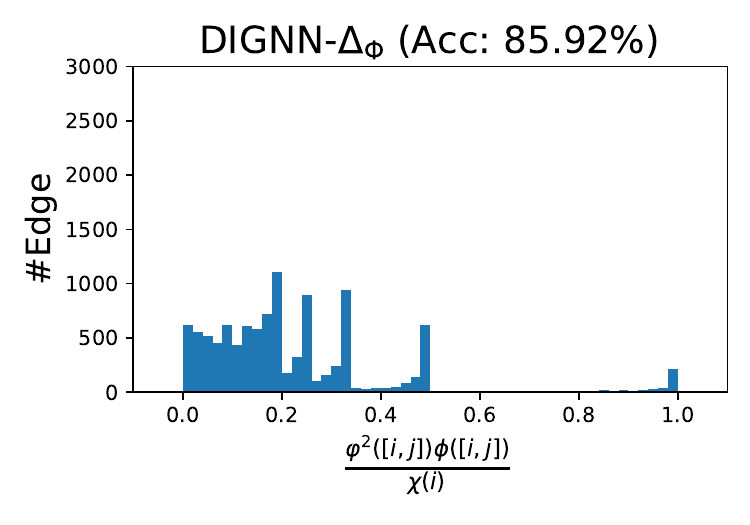}
        \caption{Learned $\chi, \phi,\varphi$ for Cora.}\label{app:fig:cora-learn}
    \end{subfigure}
    \begin{subfigure}[b]{0.4\textwidth}
        \centering
        \includegraphics[width=\textwidth]{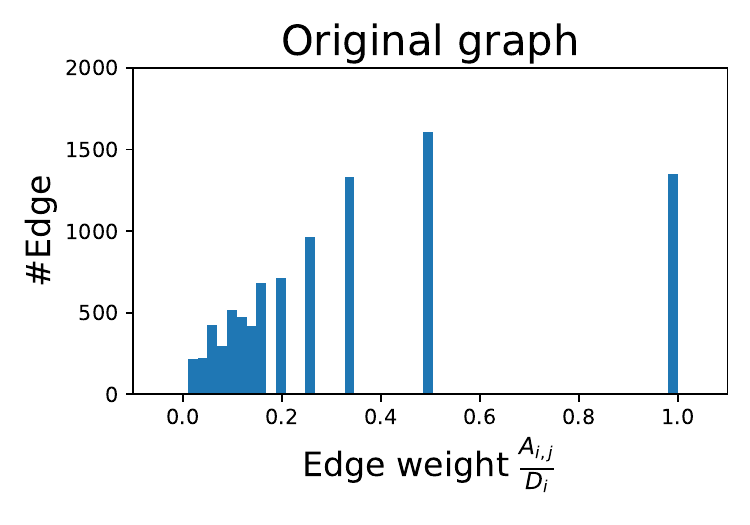}
        \caption{Original edge weight of CiteSeer.}\label{app:fig:citeseer-org}
    \end{subfigure}
    \begin{subfigure}[b]{0.4\textwidth}
        \centering
        \includegraphics[width=\textwidth]{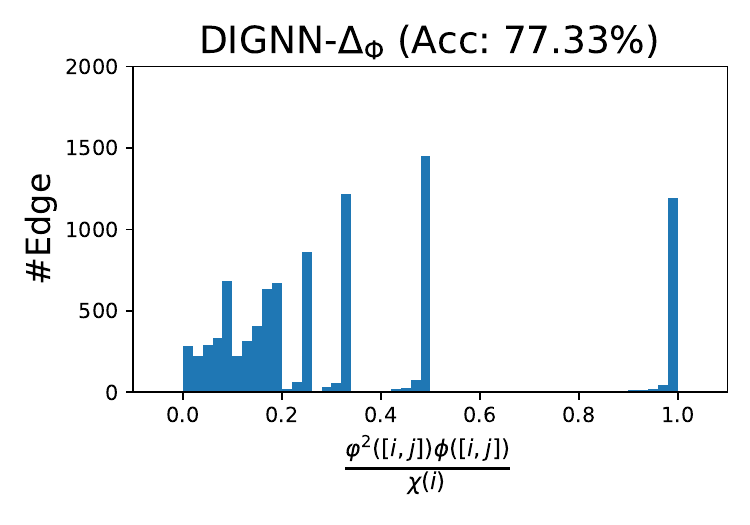}
        \caption{Learned $\chi, \phi,\varphi$ for CiteSeer.}\label{app:fig:citeseer-learn}
    \end{subfigure}
    \begin{subfigure}[b]{0.4\textwidth}
        \centering
        \includegraphics[width=\textwidth]{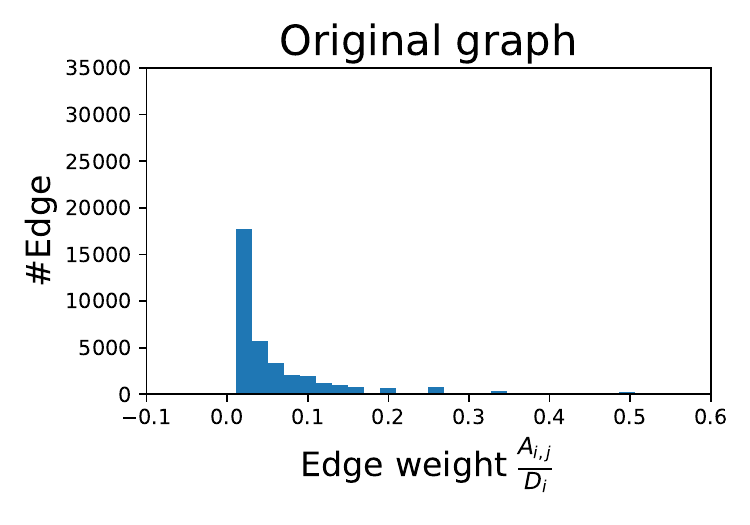}
        \caption{Original edge weight of Chameleon.}\label{app:fig:chameleon-org}
    \end{subfigure}
    \begin{subfigure}[b]{0.4\textwidth}
        \centering
        \includegraphics[width=\textwidth]{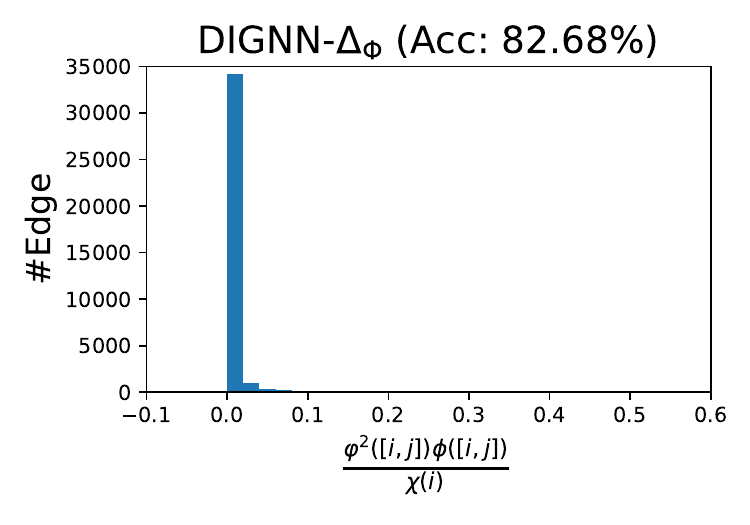}
        \vspace{-5pt}
        \caption{Learned $\chi, \phi,\varphi$ for Chameleon.}\label{app:fig:chameleon-learn}
    \end{subfigure}
    \begin{subfigure}[b]{0.4\textwidth}
        \centering
        \includegraphics[width=\textwidth]{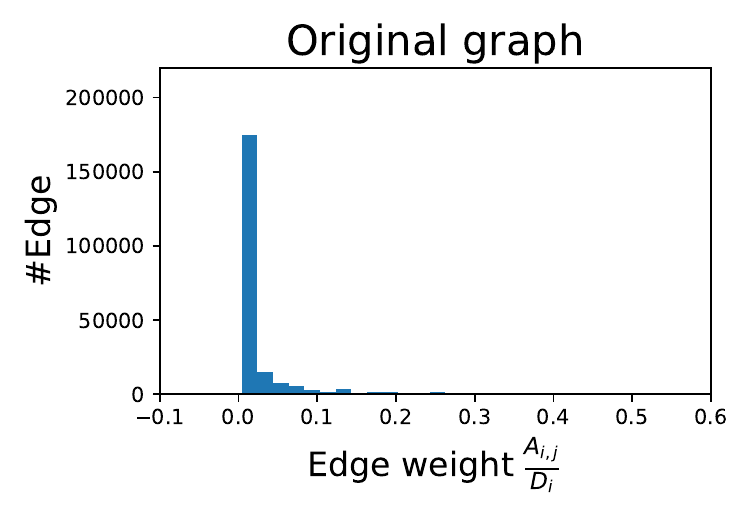}
        \caption{Original edge weight of Squirrel.}\label{app:fig:squirrel-org}
    \end{subfigure}
    \begin{subfigure}[b]{0.4\textwidth}
        \centering
        \includegraphics[width=\textwidth]{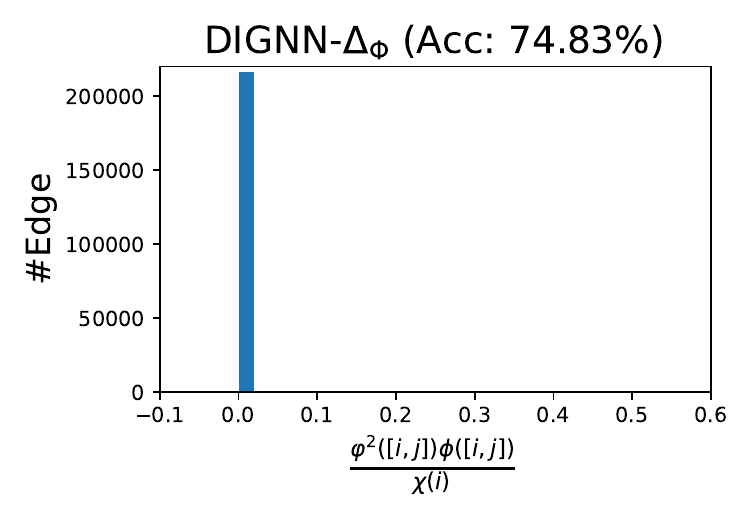}
        \caption{Learned $\chi, \phi,\varphi$ for Squirrel.}\label{app:fig:squirrel-learn}
    \end{subfigure}
    \caption{Results for the learned graph metric.}\label{app:fig:func}
\end{figure}

\clearpage

\subsection{Results for the learned positive real-valued functions}
Here, we compare the values induced by the learned positive real-valued functions and the values induced by the original edge weights for several heterophilic and homophilic benchmark datasets. 
We present the distributions of the values of $\frac{\varphi^2([i, j])\phi([i, j])}{\chi(i)}$ of DIGNN-$\Delta_\Phi$ over all edges and the distribution of original edge weights in \cref{app:fig:func}.

We observe from \cref{app:fig:cora-org,app:fig:cora-learn,app:fig:citeseer-org,app:fig:citeseer-learn} that the distributions of the values of $\frac{\varphi^2([i, j])\phi([i, j])}{\chi(i)}$ on homophilic benchmark datasets, i.e., Cora and CiteSeer, are similar to the distributions of original normalized edge weights, which indicates that aggregating and transforming node features over the original graph topology is highly beneficial for label prediction. It explains why our models and explicit GNN baselines archive similar performance on these homophilic benchmark datasets.

Contradict to the results of \cref{app:fig:chameleon-org,app:fig:chameleon-learn,app:fig:squirrel-org,app:fig:squirrel-learn} on homophilic graphs, the results on heterophilic graphs, i.e., Chameleon and Squirrel, show that the distributions of the values of $\frac{\varphi^2([i, j])\phi([i, j])}{\chi(i)}$ are substantially different from the distributions of the original edge weights. We observe that the values learned by DIGNN-$\Delta_\Phi$ are mostly close to zero for most edges, indicating that the propagation weights of DIGNN-$\Delta_\Phi$ are generally very small. It illustrates that the local neighborhood information in Chameleon and Squirrel is not as helpful for label prediction as it is in Cora and CiteSeer. Consequently, propagating and transforming node features over the original graph topology may hinder the performance of GNNs on these heterophilic datasets.

\clearpage

\section{Computational Complexity and Running Time of DIGNNs}\label{app:complex}

\subsection{Computational complexity of the implicit graph diffusion layer in DIGNN-$\Delta_\Phi$}
Given a graph $\gG = (\gV, \gE, \mA)$ with $n$ vertices and $m$ edges (suppose that $m > n$), node embeddings $\widetilde{\mX} = (\widetilde{\vx}_1, \dots, \widetilde{\vx}_n)^\top \in \R^{n \times d}$. Let parameters $\Theta_\chi \in \R^{h \times d}, \Theta_\phi \in \R^{h \times h}, \Theta_\varphi \in \R^{h \times d}$ and the positive real-valued functions $\chi, \phi, \varphi$ are defined as \cref{eq:eq12,eq:eq13,eq:eq14} respectively, i.e., $\forall i \in \gV$ and $\forall [i, j] \in \gE$, 
\begin{align*}
    \chi(i) = {} & \emD_i\tanh(\|\Theta_\chi\widetilde{\vx}_i\|), \\
    \phi([i, j]) = {} & \tanh(|(\Theta_\phi\Theta_\chi\widetilde{\vx}_i)^\top(\Theta_\phi\Theta_\chi\widetilde{\vx}_j)|), \\
    \varphi([i, j]) = {} & \sqrt{\emA_{i,j}\tanh((\|\Theta_\varphi(\widetilde{\vx}_i - \widetilde{\vx}_j)\| + \epsilon)^{-1})}, 
\end{align*}
Note that by \cref{eq:param-lap}:
\begin{equation*}
    (\Delta_\Phi f)(i) = \frac{1}{\chi(i)}\sum_{j=1}^n\varphi([i, j])^2\phi([i, j])\left(f(i) - f(j)\right).
\end{equation*}
for $\mZ^{(t)} = (\vz_1^{(t)}, \dots, \vz_n^{(t)}), t = 0, 1, \dots, T$, we obtain
\begin{equation*}
    \left(\Delta_\Phi\mZ^{(t)}\right)_{i,:} = \frac{1}{\chi(i)}\sum_{j=1}^n\varphi([i, j])^2\phi([i, j])\left(\vz_i^{(t)} - \vz_j^{(t)}\right).
\end{equation*}
Consider the iteration equation~\cref{eq:iter} with $\Delta \equiv \Delta_\Phi$ that is used to solve the implicit GNN layer of DIGNN-$\Delta_\Phi$, i.e.,
\begin{equation*}
    \mZ^{(t+1)} = \widetilde{\mX} - \frac{1}{\mu}\Delta_\Phi\mZ^{(t)}, \text{ with } \mZ^{(0)} = \mathbf{0}, t = 0, 1, \dots, T,
\end{equation*}
we break down its computational complexity analysis as follows:
\begin{itemize}
    \item  \textbf{Computing \(\chi(i)\):} This involves a matrix-vector multiplication $\Theta_\chi\widetilde{\vx}_i$ with complexity $O(hd)$, followed by a norm ($O(h)$) and an application of $\tanh$ ($O(h)$). The total complexity for computing $\chi(i)$ is $O(hd + h) = O(hd)$.

    \item  \textbf{Computing $\phi([i, j])$:} This requires a matrix-vector multiplication $\Theta_\phi\Theta_\chi\widetilde{\vx}_i$ with complexity $O(h^2 + hd)$. Similarly for $\Theta_\phi\Theta_\chi\widetilde{\vx}_i$ we we need $O(h^2 + hd)$ and $(\Theta_\phi\Theta_\chi\widetilde{\vx}_i)^\top(\Theta_\phi\Theta_\chi\widetilde{\vx}_j)$ is $O(2(h^2 + hd) + h^2) = O(h^2 + hd)$. The absolute value and $\tanh$ are $O(1)$ operations. The total complexity for computing $\phi([i, j])$ is \(O(h^2 + hd)\).

    \item \textbf{Computing $\varphi([i, j])$:} Similar to $\phi([i, j])$, but involves vector subtraction $O(d)$, a matrix-vector multiplication $\Theta_\varphi(\widetilde{\vx}_i - \widetilde{\vx}_j)$ with complexity $O(hd)$, and a norm plus element-wise operations. Thus, the complexity is $O(hd)$.

    \item \textbf{Computing $\Delta_\Phi\mZ^{(t)}$:} The dominant term comes from computing $\phi([i, j])$ for all $m$ edges, leading to $O(m(h^2 + hd))$.

    \item \textbf{Updating \(\mZ^{(t+1)}\):} The update equation involves subtracting a matrix from another $O(n \times d)$ and scaling by $1/\mu$ with $O(n \times d)$, which is negligible compared to the complexity of computing $\Delta_\Phi\mZ^{(t)}$.
\end{itemize}
Therefore, the total computational complexity per iteration is $O(m(h^2 + hd))$. Since this update is repeated for $T$ iterations, the overall computational complexity of \cref{eq:iter} is $O(Tm(h^2 + hd))$.

\subsection{Running time of DIGNNs}
\cref{App:tab:time_homo} reports the averaged running time of  DIGNN-$\Delta^{\text{(rw)}}$ and DIGNN-$\Delta_\Phi$ on several benchmark datasets. Our experiments are conducted on a server equipped with a GeForce RTX 2080 Ti and 12GB of GPU memory.
\begin{table}[htp]
    \centering
    \caption{Running of DIGNNs on node classification datasets. Averaged running time per epoch (ms) / averaged total running time (s).}\label{App:tab:time_homo}
    \resizebox{\linewidth}{!}%
    {
    \begin{tabular}{lccccc}
        \toprule
        Method & Chameleon & Squirrel & Cora & CiteSeer & PubMed \\
        \midrule
        DIGNN-$\Delta^{\text{(rw)}}$ & 22.38ms / 32.78s & 60.58ms / 104.25s & 22.02ms / 30.35s & 23.79ms / 36.08s & 34.82ms / 51.21s \\
        DIGNN-$\Delta_\Phi$ & 38.89ms / 67.74s & 133.46ms / 249.48s & 28.92ms / 46.38s & 30.95ms / 40.62s & 71.11ms / 123.61s \\
        \bottomrule
    \end{tabular}
    }
\end{table}


\end{document}